\documentclass[10pt,twocolumn,letterpaper]{article}
\usepackage[pagenumbers]{wacv}

\usepackage[T1]{fontenc}

\usepackage{dsfont}

\definecolor{wacvblue}{rgb}{0.21,0.49,0.74}
\usepackage[pagebackref,breaklinks,colorlinks,allcolors=wacvblue]{hyperref}

\def\confName{WACV}
\def\confYear{2027}

\newcommand{\sys}{BodyBridge}

\title{\sys: Grounded Body Layers for Sim-to-Real Translation of
Vulnerable Road Users}

\author{Mark Azer \qquad Fabian B.~Flohr\\
\small Intelligent Vehicles Lab\\
Munich University of Applied Sciences\\
{\tt\small \{mark.azer,fabian.flohr\}@hm.edu}}

\hypersetup{
  pdftitle={BodyBridge: Grounded Body Layers for Sim-to-Real Translation of Vulnerable Road Users},
  pdfauthor={Mark Azer; Fabian B. Flohr}
}

\begin{document}
\twocolumn[{%
\renewcommand\twocolumn[1][]{##1}%
\maketitle
\begin{center}
  \captionsetup{type=figure}
  \includegraphics[width=\textwidth]{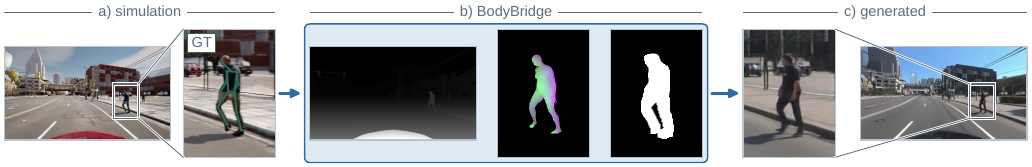}
  \captionof{figure}{\textbf{BodyBridge: Preserving articulated VRU state during
sim-to-real video generation.}
(a) Simulation provides the exact scene and articulated VRU state.
(b) BodyBridge converts this state into Grounded Body Layers for
scene- and body-aware control.
(c) The generated video preserves the simulator-specified body
configuration and facing direction.}
  \label{fig:teaser}
\end{center}%
}]
\begin{abstract}
Simulation-to-real driving  video diffusion models can translate photorealistic scenes, but they typically represent pedestrians and cyclists, vulnerable road users (VRUs), as generic object classes. This is insufficient for vulnerable road users whose safety-relevant appearance depends on articulated pose and identity. We present BodyBridge and Grounded Body Layers (GBL), a simulator-derived, pixel-aligned representation that combines scene geometry, canonical per-pixel body correspondence, and exact VRU masks. GBL provides explicit articulation to the video diffusion model, emphasizes VRU regions during control-branch post-training, and supports additive composition with scene-geometry control at inference. It also enables per-joint evaluation of generated VRUs and a failure taxonomy for identity errors. On CARLA-to-real translation, BodyBridge improves near-range PCK@0.5 by $22\%$ over the released depth-conditioned baseline.  Replacing normal-map body control with canonical correspondence reduces body flips by $18\%$, while composition reduces near-range facing flips by $27\%$. These results show that explicit body grounding complements scene-level control and improves pose-faithful VRU generation. All code, data, and model checkpoints will be made publicly available to support reproducible, pedestrian-aware autonomous driving research.
\end{abstract}

\section{Introduction}
\label{sec:intro}

Simulation is attractive for autonomous-driving data generation because it provides exact scene state and controllable behavior, but its rendered appearance differs substantially from real video. Video diffusion models can reduce this appearance gap, yet their conditioning is usually defined at the scene or object level, for example through depth, semantic layouts, or 3D boxes. Such representations constrain rigid traffic participants comparatively well because a vehicle's geometry does not change with articulation. For pedestrians and cyclists, however, a correct box or depth silhouette does not uniquely determine limb configuration, left-right identity, or facing direction. As a result, a translation can remain visually plausible at the scene level while changing safety-relevant VRU pose. The same limitation extends to evaluation. Driving-video generation is commonly assessed with scene-level realism or distribution metrics, whereas fine-grained VRU articulation is rarely measured against exact ground truth. In real video, body pose must itself be estimated, so the reference contains reconstruction error. A simulator offers a stronger setting: the exact body state that conditions generation can also define a separate geometric ground-truth reference for evaluation. This enables evaluation not only of whether a generated pedestrian appears realistic, but also of whether the generated body preserves the intended articulation.

We address this problem with BodyBridge, a body-grounded conditioning framework for sim-to-real video diffusion. We introduce Grounded Body Layers (GBL), a representation comprising three pixel-aligned signals derived from the same simulator state: scene geometry, dense canonical body correspondence, and an exact VRU mask. As illustrated in \cref{fig:teaser}, these signals provide complementary information: geometry preserves scene structure, canonical correspondence anchors body-surface identity across articulation, and the VRU mask identifies the image regions on which training should place increased emphasis.

Our contributions are threefold. (1) We introduce Grounded Body Layers (GBL), a simulator-derived representation that explicitly incorporates articulated-body structure into scene-conditioned sim-to-real video generation. (2) We present BodyBridge, a conditioning framework that combines body-aware control, VRU-focused training, and geometry--body composition for pose-faithful video translation, improving near-range PCK over the scene-geometry baseline. (3) We introduce, to our knowledge, the first evaluation framework for sim-to-real driving video that measures VRU articulation against exact simulator ground truth, covering pose fidelity, recovery, and identity failures. We further show that pose fidelity is strongly associated with downstream pedestrian-detection performance.

\section{Related Work}
\label{sec:related}

\textbf{Controllable Driving Video Generation:} Driving-video generation methods differ in the granularity of their conditioning. Methods such as MagicDrive, Panacea, DriveDreamer, and VISTA condition generation on semantic layouts, depth, camera geometry, trajectories, or 3D bounding boxes~\citep{magicdrive,magicdrive3d,magicdrivev2,panacea,panaceaplus,drivedreamer,drivedreamer2,gaia1,gaia2,vista,xworld,geodrive,genesis,drivephysica}. These signals constrain scene structure, viewpoint, object extent, and coarse geometry, specifying where traffic participants appear and how the surrounding scene is organized. However, such controls do not uniquely determine the articulated configuration of a vulnerable road user. Different limb configurations, left-right identities, and facing directions can remain compatible with the same bounding box or a similar depth silhouette.

Simulator-conditioned methods derive structural controls directly from simulator state. SimGen, Dreamland, DriveCtrl, Driving with DINO, and RealityBridge use simulator-derived geometry, semantics, trajectories, or object annotations to guide photorealistic generation~\citep{simgen,dreamland2025,drivectrl2026,drivingdino2026,realitybridge2026}. Cosmos-Transfer likewise supports multiple spatial control signals~\citep{cosmostransfer1,cosmostransfer25}, while its autonomous-driving formulation represents traffic participants primarily through object-level cues such as 3D boxes~\citep{cosmosdrivedreams}. These approaches provide scene- and object-level control from simulator state, but do not explicitly specify body articulation.

Finer human control has also been explored. GEM conditions generation on estimated human pose together with scene and ego-motion information~\citep{gem}, while pedestrian video editing uses reconstructed 2D pose to control individual pedestrians~\citep{pedvideoedit}. These methods explicitly constrain articulation, but their pose representations are sparse and estimated from real video. BodyBridge instead uses simulator-exact articulated state as body-level conditioning, constraining VRU articulation beyond scene geometry and object extent.

\textbf{Articulated Human Representation:} Human-video generation represents articulated pose at different levels of detail. Sparse 2D skeletons directly encode joint locations and are widely used in pose-driven animation~\citep{animateanyone,mimicmotion,human4dit,mimo,realismotion}. They are simple and interpretable, but do not represent the body surface or persistent body-part identity. Denser geometric representations use SMPL depth or normal renders~\citep{champ} and other SMPL-based body renders~\citep{realisdancedit2025}. These representations constrain body geometry more strongly, but depth and surface orientation do not uniquely identify anatomical location under left-right or front-back symmetries. DensePose IUV maps instead associate image pixels with locations on a canonical body surface~\citep{densepose,magicanimate}, providing persistent surface correspondence across articulation rather than local geometry alone. BodyBridge builds on this correspondence principle using exact body state available from simulation, while scene geometry remains a separate source of structural information.

\textbf{VRU Augmentation and Evaluation:} Generative driving data are evaluated through both downstream utility and generation quality. Augmentation studies measure whether synthetic data improve perception~\citep{veria,diffpeddet,pedvideoedit}, while generation-focused work measures realism, identity consistency, motion quality, or pedestrian dynamics~\citep{worldlens,consisdrive,pedra2025,humanscore2026,cdfvd}. These evaluations are complementary, but do not directly test whether a generated VRU preserves the articulation provided by its source state. This distinction matters because improved visual realism can still coincide with deviation from known body ground truth~\citep{genbedlam2024}. Pose-grounded evaluation has been demonstrated for sim-to-real hands~\citep{pam}, whereas driving datasets provide only reconstructed pedestrian skeletons~\citep{waymo3dskelmo}. Simulation instead exposes the articulated state directly, enabling per-joint evaluation against exact VRU ground truth.
\begin{figure*}[t]
  \centering
  \includegraphics[width=\textwidth]{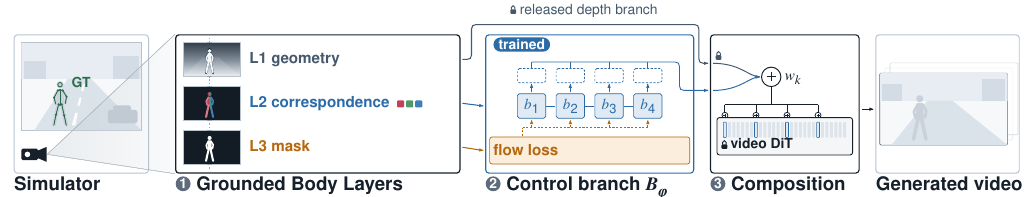}
  \caption{\textbf{BodyBridge: Grounded body control for sim-to-real
video generation.}
(1) Grounded Body Layers encode scene geometry, canonical body
correspondence, and the VRU mask from simulation.
(2) The correspondence layer, encoded in RGB, trains a dedicated body-control branch,
with the mask weighting the training objective.
(3) At inference, body and released geometry control are composed to
generate the translated video.
Green denotes evaluation ground truth only, amber the objective and its
weighting, and blue BodyBridge-specific and trained components;
padlocks indicate frozen modules.}
  \label{fig:pipeline}
\end{figure*}
\section{Methodology}
\label{sec:method}

 \cref{fig:pipeline} summarizes BodyBridge. Each recorded simulator sequence provides scene geometry, exact articulated VRU state, instance masks, and camera calibration. In Stage 1, these signals are transformed into three pixel-aligned Grounded Body Layers (GBL): scene geometry, canonical body correspondence, and a VRU mask. In Stage 2, the correspondence layer provides body-level conditioning to a dedicated control branch, while the VRU mask spatially weights the training objective; the video-generation backbone remains frozen. In Stage 3, the trained body branch is composed with scene-geometry control through an additive multi-control interface to generate the translated video. The recorded simulator skeleton also provides the reference pose used for evaluation, but remains outside the training signals. The following sections describe the three generation stages and the ground-truth evaluation protocol.

\subsection{Stage 1: Grounded Body Layer Construction}
\label{sec:method-gbl}
Stage 1 transforms each recorded simulator frame into three
pixel-aligned Grounded Body Layers (GBL), all derived from the same
articulated state and camera calibration. \textbf{$L_1$ geometry} is
metric scene depth and provides scene-level structural control.
$L_2$ correspondence assigns visible VRU pixels a canonical
body-correspondence encoding and provides body-level conditioning.
$L_3$ mask is the simulator-provided VRU instance mask and
later weights the Stage~2 training objective. $L_1$ and $L_3$ are
obtained directly from simulator annotations, whereas constructing
$L_2$ requires registering a canonical body template to the recorded
VRU articulation.

For VRU track $k$ at frame $t$, let $\theta_{k,t}$ denote the SMPL
pose parameters, $\beta_k$ the SMPL body-shape parameters, and
$\{\mathbf{j}_{k,t,m}\}_{m=1}^{M}$ the corresponding recorded
simulator joints in camera coordinates. We use $M=12$ limb
correspondences and pose a fixed $\beta=0$ SMPL template, yielding
template joints $\hat{\mathbf{j}}_m(\theta_{k,t})$ and vertices
$\hat{\mathbf{v}}_i(\theta_{k,t})$. Because the simulator actor and
SMPL template need not share the same body proportions, we align them
with a per-frame similarity transform,
\begin{equation}
(s,\mathbf{R},\boldsymbol{\tau})_{k,t}
=
\arg\min_{s,\mathbf{R},\boldsymbol{\tau}}
\sum_{m=1}^{M}
\left\|
s\mathbf{R}\hat{\mathbf{j}}_m(\theta_{k,t})
+\boldsymbol{\tau}
-\mathbf{j}_{k,t,m}
\right\|_2^2 ,
\label{eq:registration}
\end{equation}
where $s>0$, $\mathbf{R}\in SO(3)$, and
$\boldsymbol{\tau}\in\mathbb{R}^3$. We solve
\cref{eq:registration} in closed form using the Umeyama algorithm.
The registered mesh vertices are
\begin{equation}
\mathbf{v}^{\mathrm{reg}}_{k,t,i}
=
s_{k,t}\mathbf{R}_{k,t}
\hat{\mathbf{v}}_i(\theta_{k,t})
+\boldsymbol{\tau}_{k,t}.
\label{eq:registered-vertices}
\end{equation}
Meshes from all VRUs in the frame are projected through the stored
camera and jointly z-buffered, so only the front-most body surface
contributes at each image pixel.

Geometry alone does not assign persistent anatomical identity to
visually symmetric body surfaces. We therefore construct $L_2$ from
coordinates attached to the fixed canonical body template rather than
from local surface orientation.  Let
$\tilde{\mathbf{v}}_i\in\mathbb{R}^{3}$ denote the rest-pose
coordinate of template vertex $i$. For a pixel $p$ covered by the
front-most registered mesh of track $k$, let $(v_1,v_2,v_3)$ denote
the visible triangle and $(\alpha_1,\alpha_2,\alpha_3)$ its
barycentric weights, with $\sum_{i=1}^{3}\alpha_i=1$. The
correspondence channels are
\begin{equation}
\begin{aligned}
L_2^{R}(p)
&=
\phi_x\!\left(
\sum_{i=1}^{3}\alpha_i\tilde v_i^{(x)}
\right),
&
L_2^{G}(p)
&=
\phi_z\!\left(
\sum_{i=1}^{3}\alpha_i\tilde v_i^{(z)}
\right),
\\[2pt]
L_2^{B}(p)
&=
\frac{
d_{\max}^{k,t}-d_t(p)
}{
d_{\max}^{k,t}-d_{\min}^{k,t}
}.
\end{aligned}
\label{eq:ccdo}
\end{equation}
where $d_t(p)$ is the camera depth of the rendered body surface and
$d_{\min}^{k,t}$ and $d_{\max}^{k,t}$ denote the depth extrema of the
visible registered body for track $k$ at frame $t$. For canonical
axis $a$,
\begin{equation}
\phi_a(q)
=
\frac{
q-\tilde v_{\min}^{(a)}
}{
\tilde v_{\max}^{(a)}-\tilde v_{\min}^{(a)}
},
\label{eq:canonical-normalization}
\end{equation}
where $\tilde v_{\min}^{(a)}$ and $\tilde v_{\max}^{(a)}$ are the
minimum and maximum template coordinates along axis $a$. Canonical
$x$ encodes left-right body location and canonical $z$ front-back
location, while the third channel encodes intra-body depth order
independently of absolute scene distance.

The first two channels preserve canonical surface correspondence
across articulation, while the third distinguishes visible surface
ordering when body parts overlap in projection. Because the canonical
attributes are defined on the fixed rest-pose template, their values
are independent of body shape $\beta$; articulation affects only the
posed and registered location at which those attributes are rendered.
Pixels not covered by a valid VRU mesh are assigned the background
value. When mesh registration is unavailable, the affected instance
falls back to the 2D-skeleton control representation; this changes only the conditioning signal and not the evaluation reference. All three GBL layers share the same stored camera and image coordinates and are therefore pixel-aligned by construction. Registration quality and conditioning-geometry alignment are quantified in Supplementary Sec.~S2.7.

\subsection{Stage 2: Region-Weighted Post-Training}
\label{sec:method-train}
Stage~2 trains a dedicated body-control branch that maps the
$L_2$ correspondence layer to body-aware feature corrections for a
frozen video diffusion model. We instantiate this stage with
Cosmos-Transfer2.5-2B~\citep{cosmostransfer25}, whose control interface
injects features from a dedicated conditioning branch into intermediate
layers of the video DiT. Let the frozen backbone be
$F_{\psi}=f_N\circ\cdots\circ f_1$, operating on video latents
$\mathbf{z}=\mathcal{E}(\mathbf{x})\in
\mathbb{R}^{T'\times h\times w\times C}$. The body-control branch
$B_{\phi}$ encodes the correspondence video with the same latent
representation, $\mathbf{c}_0=\mathcal{E}(L_2)$, and projects it onto
the backbone token grid. The resulting conditioning tokens share the
backbone's spatiotemporal tokenization and positional encoding and
are coupled to the current denoising state at the branch input,
\[
    \mathbf{c}=P_{\mathrm{in}}(\mathbf{c}_0)+\mathbf{z}_t .
\]

The branch contains a sequence of control blocks $\{b_m\}$ that
produce spatial feature corrections at selected backbone depths. For
control block $m$, the injected feature is
$\mathbf{h}_m=P_{\mathrm{out}}^{(m)}(b_m(\mathbf{c}))$. More generally,
for $K$ simultaneously active control branches, the backbone
activation at injection location $i$ is
\begin{equation}
    \mathbf{a}_i
    \leftarrow
    \mathbf{a}_i+
    \sum_{k=1}^{K}
    w_k\,\mathbf{h}^{(k)}_{m(i)},
    \qquad i\in\mathcal{I},
    \label{eq:control-injection}
\end{equation}
where $w_k$ controls the contribution of branch $k$ and $m(i)$ maps
each injection location to its corresponding control block. Stage 2
uses only the body-correspondence branch ($K=1$). The conditioning
embedder, control blocks, and output projections constitute
$\phi$ and are post-trained, while the video backbone parameters
$\psi$ remain frozen.

Because VRUs typically occupy only a small portion of each frame,
uniform flow matching assigns comparatively little weight to their
articulated regions. The $L_3$ mask therefore enters Stage~2 as a
spatial weight on the training objective rather than as an additional
conditioning input. Let $\mathbf{z}_0$ denote the clean video latent,
$\boldsymbol{\epsilon}\sim\mathcal{N}(0,I)$, and
\[
    \mathbf{z}_t=(1-t)\mathbf{z}_0+t\boldsymbol{\epsilon},
    \qquad
    \mathbf{v}=\boldsymbol{\epsilon}-\mathbf{z}_0
\]
the rectified-flow interpolation and target velocity. We downsample
$L_3$ to the latent grid by max pooling, yielding
$\tilde{\mathbf{m}}_{\mathrm{VRU}}$, and optimize
\begin{equation}
    \mathcal{L}_{\mathrm{flow}}
    =
    \mathbb{E}_{t,\boldsymbol{\epsilon}}
    \left[
    \frac{1}{|\Omega|}
    \sum_{q\in\Omega}
    \left(1+\lambda\tilde m_{\mathrm{VRU},q}\right)
    \ell_q
    \right].
    \label{eq:flow}
\end{equation}
such that $\ell_q=|\mathbf v_q-\mathbf v_{\phi,q}(\mathbf z_t,t,L_2)|_2^2$ denotes the flow-matching error at latent location $q$ and $\Omega$ denotes the spatiotemporal latent grid. Max pooling preserves the increased weight of small VRUs that would otherwise disappear when mapped to the latent
resolution. Setting $\lambda=0$ recovers the unweighted objective.
After post-training, $B_{\phi}$ provides a dedicated body-level control
signal; Stage~3 combines this signal with scene-geometry control through
the same additive interface in \cref{eq:control-injection}.

\subsection{Stage 3: Geometry-Body Composition}
\label{sec:method-compose}
Stage 2 produces a control branch specialized for articulated VRU
structure, while scene geometry provides complementary constraints on
the surrounding environment and object layout. Stage 3 therefore
combines the two controls rather than replacing the scene-level signal
with body conditioning. We use the released depth-control branch for
$L_1$ geometry and the post-trained correspondence branch for $L_2$.
Both branches inject features through the same control
interface.

At inference, the backbone activation at each injection location is
updated by the weighted sum of the two branch outputs,
\begin{equation}
    \mathbf{a}_i
    \leftarrow
    \mathbf{a}_i
    +
    w_{\mathrm{geom}}\,
    \mathbf{h}^{\mathrm{geom}}_{m(i)}
    +
    w_{\mathrm{body}}\,
    \mathbf{h}^{\mathrm{body}}_{m(i)},
    \qquad i\in\mathcal{I}.
    \label{eq:composition}
\end{equation}
Here, $\mathbf{h}^{\mathrm{geom}}_{m(i)}$ and
$\mathbf{h}^{\mathrm{body}}_{m(i)}$ denote the feature corrections
produced by the geometry and body branches, respectively, and
$w_{\mathrm{geom}}$ and $w_{\mathrm{body}}$ control their relative
contributions.

The two branches remain independently parameterized and are combined
only through additive feature injection; no additional fusion module
is trained. The generator therefore receives scene-level
geometry from $L_1$ and articulated-body information from $L_2$
within the same denoising process.

\subsection{Simulator-Grounded VRU Evaluation}
\label{sec:method-eval}
Simulation provides an articulated reference for every generated frame,
allowing pose preservation to be measured against the source state
rather than against a pose reconstructed from the generated video.
We keep conditioning and evaluation representations separate: the
GBL body layer is constructed from the registered SMPL representation,
whereas evaluation uses the recorded CARLA skeleton. For each simulation frame, the simulator joints are projected through
the stored camera and mapped to COCO-17 keypoints
$\{\mathbf{g}_{n,j}\}$, where $n$ indexes a VRU instance and $j$
indexes a keypoint. A held-out pose estimator predicts the
corresponding keypoints $\{\hat{\mathbf{g}}_{n,j}\}$ from the
generated frame.

Let $\mathcal{S}$ denote the set of scorable ground-truth VRU
instances and $\mathcal{V}_n$ the set of valid ground-truth keypoints
for instance $n\in\mathcal{S}$. We define its torso normalization
scale as
\begin{equation}
d_n =
\left\|
\mathbf{g}_{n,\mathrm{L.shoulder}}
-
\mathbf{g}_{n,\mathrm{R.hip}}
\right\|_2 .
\label{eq:torso-scale}
\end{equation}
Let $r_n\in\{0,1\}$ indicate whether the pose estimator recovers
instance $n$, and let
\[
e_{n,j}
=
\left\|
\hat{\mathbf{g}}_{n,j}
-
\mathbf{g}_{n,j}
\right\|_2
\]
denote the image-space joint error for a recovered instance.

For normalized distance threshold $\alpha$, Percentage of Correct
Keypoints is
\begin{equation}
\mathrm{PCK}@\alpha =
\frac{
\displaystyle
\sum_{n\in\mathcal{S}}
\sum_{j\in\mathcal{V}_n}
\mathds{1}
\!\left[
r_n=1
\land
e_{n,j}\leq\alpha d_n
\right]
}{
\displaystyle
\sum_{n\in\mathcal{S}}
|\mathcal{V}_n|
},
\label{eq:pck}
\end{equation}
where $\mathrm{1}$$[\cdot]$ is the indicator function. Thus, all valid
keypoints of a missed instance contribute zero to PCK.

We report instance recovery separately as
\begin{equation}
\mathrm{Recovery}
=
\frac{1}{|\mathcal{S}|}
\sum_{n\in\mathcal{S}} r_n .
\label{eq:recovery}
\end{equation}

For recovered instances, normalized median joint error is
\begin{equation}
\mathrm{NMJE}
=
\operatorname*{median}_{\substack{
n\in\mathcal{S}:r_n=1\\
j\in\mathcal{V}_n}}
\frac{e_{n,j}}{d_n}.
\label{eq:nmje}
\end{equation}
To prevent distorted or missing VRUs from disappearing from the
evaluation population, we use a top-down protocol defined by the
simulator ground-truth boxes. The pose estimator is applied to every
scorable GT instance, irrespective of whether that instance would be
detected in a bottom-up pass. This fixes the GT denominator across
generation variants and separates loss of recovery from joint
localization error.

PCK captures the magnitude of pose error but not its type. We
therefore further categorize failures into \emph{miss},
\emph{full-body flip}, \emph{switch}, \emph{merge},
\emph{stump}, and \emph{residual} errors, and assess facing direction
separately from the projected shoulder ordering. The operational
criteria and thresholds for these categories are specified in
Supplementary Sec.~S1.1.
\begin{figure*}[t]
  \centering
  \includegraphics[width=\linewidth]{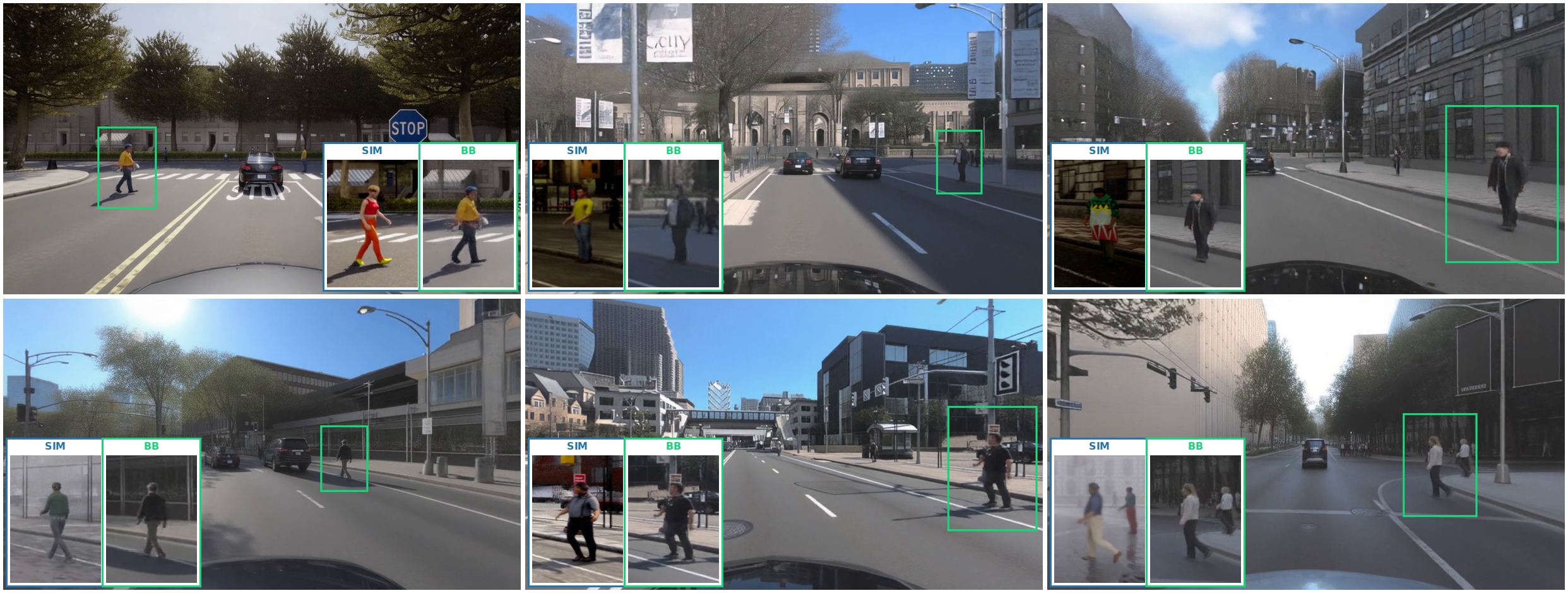}
  \caption{\textbf{Qualitative sim-to-real results with BodyBridge.}
Examples across diverse scenes and pedestrian configurations show
appearance translation while retaining the simulator-specified VRU
pose and scene placement.
Green boxes mark the highlighted VRUs; insets compare the source
simulator crop (\textsc{Sim}) with the corresponding BodyBridge
generation (\textsc{BB}).}
  \label{fig:qual}
\end{figure*}
\section{Experimental Setup}
\label{sec:setup}

\subsection{Data and Splits}
\label{sec:exp-data}

We instantiate BodyBridge on HABIT~\citep{habit} sequences rendered in
CARLA. Each frame provides articulated VRU state, instance masks,
metric depth, object annotations, and camera calibration. We export
$93$-frame clips at $16$\,fps and $1280\times720$ resolution.
SMPL parameters construct the body-control representation, while the
recorded CARLA skeleton provides the separate evaluation reference
of \cref{sec:method-eval}.

Training uses $1{,}650$ clips from $220$ route subsets and $3{,}470$
pedestrian motions; evaluation uses $600$ clips from $80$ held-out
routes and $1{,}260$ held-out motions, for $4{,}730$ motions in total. Each evaluation source is exported once and
reused across all variants, so comparisons share scene, motion,
camera, and GT and differ only in conditioning or training.
The top-down set contains $16{,}217$ scorable VRU instances per
variant: $2{,}531$ near ($<15$\,m), $6{,}694$ mid
($15$-$30$\,m), and $6{,}992$ far ($>30$\,m).

\subsection{Training and Inference}
\label{sec:exp-training}
We post-train the BodyBridge control branch on
Cosmos-Transfer2.5-2B~\citep{cosmostransfer25} for $2{,}200$ iterations using VRU-centered crops, while keeping the video
DiT frozen. Four control blocks inject at
$\mathcal{I}=\{0,7,14,21\}$; the blocks initialize from the
corresponding backbone blocks, while the control embedder and
output projections warm-start from the released edge-control
checkpoint. All control-branch parameters are trainable. We use
$\lambda=4$ in \cref{eq:flow}, with the VRU mask max-pooled to the
latent grid. At inference, the released depth and trained body branches
are composed with fixed weights
$(w_{\mathrm{geom}},w_{\mathrm{body}})=(0.5,1.0)$, chosen before
evaluation. All variants use the same byte-identical 33-word prompt. Additional training and implementation details are provided in
Supplementary Sec.~S5.2.

\subsection{Baselines and Ablations}
\label{sec:exp-baselines}

We organize the comparison around three factors: body representation,
Stage~2 training, and Stage~3 geometry-body composition. The released Cosmos depth-control branch serves as the scene-geometry baseline, and the released edge-control branch as an additional spatial-control reference.

For body representation, all variants use the same articulated
simulator state and compare a 2D skeleton, a rendered 3D body normal
map, and the proposed canonical correspondence layer. The 2D skeleton
provides sparse joint-level control, whereas the normal-map and
correspondence variants provide dense pixel-aligned body control.

For Stage~2 training, we compare the region-weighted objective with an unweighted variant that removes the additional VRU emphasis
($\lambda=0$), and compare released-control warm-start initialization with
base initialization. For Stage~3 composition, the normal-map and
correspondence branches are evaluated both alone and in combination
with the released depth-control branch.

All variants use the same evaluation clips, articulated source states,
camera trajectories, GT boxes, and evaluation protocol. Untranslated
simulator RGB, control-copy, and permutation variants are included as
evaluation and control-adherence references.

\subsection{Metrics and Statistical Protocol}
\label{sec:exp-metrics}

We use YOLO11-L-pose as a held-out evaluation estimator in the
top-down protocol of \cref{sec:method-eval}. An instance is scorable
when at least three GT keypoints are valid and the torso scale $d$ is
defined. We report PCK at
$\alpha\in\{0.1,0.2,0.5\}$ by distance
(near $<15$\,m, mid $15$-$30$\,m, far $>30$\,m), recovery, and
NMJE, the median normalized error over all valid keypoints of recovered
instances. Missing instances score zero in PCK.

Near PCK@0.5 is the primary pose-fidelity metric, with recovery
co-primary. Pooled PCK mixes recovery and articulation and is strongly
influenced by the far stratum, where even untranslated simulator RGB
reaches only $0.120$ PCK@0.5; pooled, mid-, and far-range PCK are
therefore reported as complementary coverage summaries. NMJE measures
joint localization conditional on recovery. Appearance is evaluated
separately using VRU-crop KID against a real KITTI pedestrian set
disjoint from real-video post-training data.

For results from a single inference seed, we report $95\%$ percentile
CIs from $10{,}000$ paired clip-level bootstrap resamples over the
$600$ evaluation clips. For key BodyBridge-depth and
canonical correspondence-normal map comparisons, we additionally regenerate each
configuration with three inference seeds and use a clip$\times$seed
bootstrap. Checkpoints are fixed, so these analyses capture clip and
inference-seed variation. Training-run robustness is assessed separately using independently
trained canonical correspondence checkpoints (Supplementary Sec.~S2.2).

\section{Results}
\label{sec:results}

\subsection{BodyBridge vs. Scene-Geometry Control}
\label{sec:results-main}
\Cref{tab:main} compares BodyBridge with the released Cosmos depth
control under the simulator-grounded top-down protocol. BodyBridge
raises the primary near-range PCK@0.5 from $0.666$ to $0.812$ and
reduces NMJE from $0.380$ to $0.282$. Pooled PCK likewise increases
from $0.308$ to $0.373$, while mid-range PCK improves from $0.407$ to
$0.513$. Across three inference seeds, the
BodyBridge-depth near-PCK difference remains positive
($+0.142$, $95\%$ CI $[+0.101,+0.183]$). The near-range ordering also persists under a disjoint pose estimator
(Supplementary Sec.~S2.6). Qualitative examples in
\cref{fig:qual} show the corresponding translations across
different scenes and pedestrian configurations, with the generated
VRUs retaining the prescribed body configuration and scene placement.

\begin{table}[ht]
\centering
\caption{\textbf{BodyBridge vs.\ scene-geometry control.}
Pose fidelity is evaluated against exact simulator ground truth; KID$_{\rm VRU}$ is reported as $\times10^3$.
}
\label{tab:main}
\small
\setlength{\tabcolsep}{3.6pt}
\begin{tabular}{lccccc}
\toprule
Method &
Recov. $\uparrow$ &
PCK$_{\rm all}$ $\uparrow$ &
Near $\uparrow$ &
NMJE $\downarrow$ &
KID $\downarrow$ \\
\midrule
Cosmos depth
& \textbf{.504}
& .308
& .666
& .380
& 217.2 \\

BodyBridge
& .493
& \textbf{.373}
& \textbf{.812}
& \textbf{.282}
& \textbf{139.3} \\
\bottomrule
\end{tabular}
\end{table}
The pose improvement is not explained by higher overall recovery:
BodyBridge recovers $0.493$ of scorable instances compared with
$0.504$ for depth. Near-range recovery nevertheless increases by
$0.037$ ($95\%$ paired CI $[+0.011,+0.070]$), while the remaining
recovery deficit is concentrated at far range.

BodyBridge also reduces VRU-crop KID from $217.2$ to $139.3$.
We treat KID as a complementary measure of appearance distribution,
rather than articulated-pose fidelity.

\subsection{Component Attribution}
\label{sec:results-attrib}

\begin{table}[ht]
\centering
\caption{\textbf{Component ablation.}
Effects of body representation, weighted post-training, and
geometry-body composition.}
\label{tab:ablation}
\small
\setlength{\tabcolsep}{3pt}
\begin{tabular}{lcccc}
\toprule
Variant &
Recov. $\uparrow$ &
PCK$_{\rm all}$ $\uparrow$ &
Near $\uparrow$ &
NMJE $\downarrow$ \\
\midrule
\multicolumn{5}{l}{\emph{Body representation}} \\
Edge (training-free)
    & .212 & .136 & .492 & .360 \\
2D skeleton
    & .320 & .184 & .595 & .461 \\
Normal map
    & .414 & .310 & .796 & .281 \\
Canonical corr.
    & \textbf{.431} & \textbf{.325} & \textbf{.805} & \textbf{.291} \\
\midrule
\multicolumn{5}{l}{\emph{Stage 2: post-training}} \\
Normal map, no weight& .390 & .279 & .738 & .332 \\
Normal map, weighted
    & \textbf{.414} & \textbf{.310} & \textbf{.796} & \textbf{.281} \\
Normal map, base init.
    & .302 & .199 & .645 & .369 \\
\midrule
\multicolumn{5}{l}{\emph{Stage 3: composition}} \\
Normal map
    & .414 & .310 & .796 & .281 \\
Depth + normal map
    & .486 & .369 & .808 & \textbf{.274} \\
Canonical corr.
    & .431 & .325 & .805 & .291 \\
Depth + Canonical corr.
    & \textbf{.493} & \textbf{.373} & \textbf{.812} & .282 \\
\bottomrule
\end{tabular}
\end{table}
\Cref{tab:ablation} isolates the three design choices of BodyBridge.
First, the conditioning representation strongly affects pose fidelity.
The 2D skeleton reaches $0.595$ near PCK@0.5, whereas the rendered
normal map reaches $0.796$ and canonical correspondence reaches
$0.805$. This gap indicates that dense, spatially aligned body
conditioning preserves articulation substantially better than sparse
joint control.

Second, region weighting improves the body-control branch. On the normal-map branch, introducing the VRU-weighted
objective raises near PCK from $0.738$ to $0.796$
($+0.058$ [$+0.027,+0.088$]), while reducing NMJE from $0.332$ to
$0.281$. Initialization also matters: under the same weighted recipe,
base initialization reaches only $0.645$ near PCK, compared with
$0.796$ for the released-control warm start. Thus, the gain from
post-training cannot be attributed to the conditioning signal alone.

Third, geometry-body composition mainly improves coverage while
preserving near-range pose fidelity. Adding depth control to the
normal-map branch increases recovery from $0.414$ to $0.486$ and
pooled PCK from $0.310$ to $0.369$, while near PCK changes from
$0.796$ to $0.808$. For canonical correspondence, recovery increases
from $0.431$ to $0.493$ and pooled PCK from $0.325$ to $0.373$, while
near PCK changes from $0.805$ to $0.812$. The aggregate difference between
the normal-map and correspondence representations is small; their
difference on the targeted identity failures is examined separately
in \cref{sec:results-taxonomy}.

\subsection{Canonical Correspondence and Identity Failures}
\label{sec:results-taxonomy}

\begin{figure}[ht]
  \centering
  \includegraphics[width=\columnwidth]{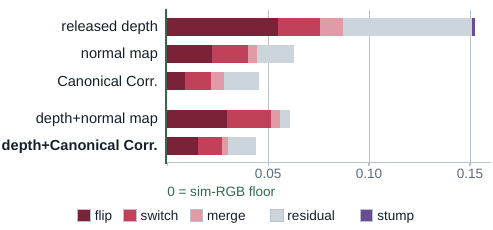}
 \caption{\textbf{Generation-induced VRU pose failures by failure type.}
Near-range failures are shown relative to the untranslated sim-RGB
estimator floor. Identity-related failures-flip, switch, and
merge-account for most of the excess errors, while canonical
correspondence reduces the flip component compared with the normal-map
representation.}
  \label{fig:taxonomy}
\end{figure}

Aggregate pose metrics only weakly separate the normal-map and canonical correspondence
representations, but the failure taxonomy reveals a more specific
difference. For the weighted normal-map branch, $60.0\%$ of failed
near-range keypoints belong to identity-related classes
(flip, switch, or merge). Because the evaluation estimator itself
produces such errors, we measure generation-induced failures relative
to the untranslated sim-RGB control using the same GT denominator.
After subtracting this estimator floor, approximately $71\%$ of the
near-range excess failures are identity-related
(\cref{fig:taxonomy}). 
Canonical correspondence targets this failure mode while leaving aggregate fidelity nearly
unchanged. Under the same weighted training recipe, near PCK increases
from $0.796$ for the normal map to $0.805$ for canonical correspondence, a difference that
is not statistically resolved. Full-body flip rate, however, decreases
from $0.108$ to $0.088$, a relative reduction of $18.4\%$
($[6.1,30.7]\%$). The same replacement inside the composed model
reduces flips by $18.2\%$ ($[5.5,29.6]\%$). 

\begin{figure}[ht]
  \centering
  \includegraphics[width=\columnwidth]{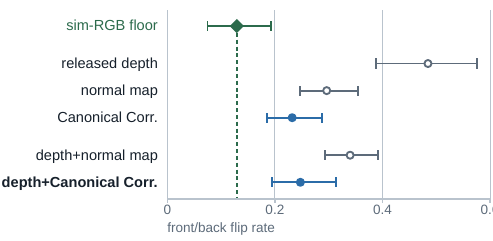}
 \caption{\textbf{Near-range VRU facing-direction errors.}
Points show front/back flip rates among recovered instances; error bars denote 95\% clip-bootstrap CIs and the green
line marks the untranslated sim-RGB estimator floor.
Canonical correspondence reduces facing flips relative to the normal
map, both alone ($-21.6\%$) and when composed with depth
($-27.2\%$).}
  \label{fig:flips}
\end{figure}

Facing errors show the same pattern. Canonical correspondence reduces near-range front/back flips by $21.6\%$ for the body
branch alone and by $27.2\%$ after composition with depth
(\cref{fig:flips}); the composed reduction is statistically
resolved. We therefore attribute the main benefit of canonical correspondence to preserving body-part and facing identity rather than to a large shift in aggregate PCK. A track-level analysis further shows that scene-geometry control can maintain an incorrect facing direction over time, while body conditioning substantially reduces these persistent errors
(Supplementary Sec.~S3.1).

\subsection{Application: Downstream VRU Detection}
\label{sec:results-downstream}
We test whether pose fidelity is associated with downstream pedestrian
detection on KITTI. For each of nine generated-data configurations, we
augment the detector training set with approximately $900$ generated
frames and train YOLO11-s for $50$ epochs with six detector seeds.
Across configurations, the added frames and GT boxes are identical;
only the generated pixel appearance differs.

Mean detection gain ranks strongly with near PCK@0.5:
Spearman $\rho=0.810$ for $\Delta$AP@0.5
(exact permutation $p=0.022$) and $\rho=0.857$ for
$\Delta$AP@[.5:.95] ($p=0.011$). The association is
stable across detector seeds, with $P(\rho>0)\geq0.997$ under a
detector-seed bootstrap.
The highest-fidelity generated source is approximately neutral for
detection, whereas lower-fidelity generated variants reduce
AP@0.5 by $0.010$-$0.039$. We therefore interpret this experiment as
a rank association between pose fidelity and downstream utility.

\subsection{Real-Data Extension}
\label{sec:results-real}

We extend body-control training with KITTI and AV2 video using
HMR2.0b~\citep{goel2023humans} pseudo-SMPL conditioning. Under the
unweighted recipe, adding real video reduces VRU-crop KID by $28\%$
without a statistically resolved pose loss and approximately halves
the downstream detection deficit.

Under the VRU-weighted objective, increasing the real-video fraction
from $0\%$ to $66\%$ and $100\%$ produces a monotonic
pose-appearance trade-off: near PCK@0.5 decreases from $0.805$ to
$0.742$ and $0.688$, while KID improves from $125.6$ to $99.6$ and
$94.0$, respectively.

These results show that real-video supervision improves appearance
alignment but, when it increasingly replaces exact simulator pose,
reduces adherence to the control body articulation. We therefore treat
pseudo-labelled real video as complementary appearance supervision,
rather than a substitute for simulator-grounded body control.

\section{Discussion and Limitations}
\label{sec:discussion}

Our evaluation measures adherence to the simulator-specified
articulation rather than perceptual realism alone. Pose-estimator priors remain a potential limitation. We mitigate
this by using an evaluation estimator that is disjoint from training,
scoring against the recorded simulator skeleton rather than the conditioning representation. 
The downstream rank association further suggests that the
pose metric captures a property relevant to pedestrian detection,
although it does not establish a causal effect.

The statistical conclusions are strongest for near- and mid-range
VRUs. Clip-level confidence intervals are complemented by regeneration
with three inference seeds for the key comparisons. BodyBridge outperforms the released depth baseline under this analysis, whereas the
aggregate pose difference between canonical correspondence and the
normal-map representation remains small. Independently trained
correspondence checkpoints show additional training-run variation, which is why we scope the correspondence claim primarily to
identity-related failures rather than aggregate PCK. Far-range pose
remains limited by the evaluation estimator and is therefore reported
mainly as a coverage summary.

Finally, the representation inherits limitations of SMPL and the
experimental setting. SMPL does not fully represent children, atypical
body shapes, clothing, carried objects, or mobility aids. Experiments
use a single simulator, camera configuration, and video-diffusion
backbone, so generalization remains to be established. Synthetic-data
pipelines may also reproduce demographic imbalance, motivating
demographic auditing before deployment.

\section{Conclusion}
\label{sec:conclusion}
We introduced BodyBridge and Grounded Body Layers (GBL) for preserving
articulated VRU state during sim-to-real driving-video generation. GBL
combines scene geometry, canonical body correspondence, and an exact
VRU mask to provide body-aware conditioning, VRU-focused post-training,
and geometry-body composition within a frozen video diffusion
backbone. We also introduced a simulator-grounded evaluation protocol
that measures pose fidelity, recovery, and identity failures against
the recorded articulated state. BodyBridge improves near-range
PCK@0.5 from $0.666$ to $0.812$ over released scene-geometry control,
while canonical correspondence reduces full-body flips by $18.4\%$
and composed facing flips by $27.2\%$. These results show that
explicit body grounding complements scene-level control when
sim-to-real generation must preserve safety-relevant VRU articulation.

\section*{Acknowledgments}
This work was supported by the German Federal Ministry for Economic
Affairs and Energy (BMWE) through the project nxtAIM (NXT GEN AI
METHODS---Generative Methods for Perception, Prediction and Planning),
under grant no.~19A23014M, based on a decision of the German Bundestag.
The authors are solely responsible for the content of this publication.

{
  \small
  \bibliographystyle{ieeenat_fullname}
  \bibliography{references}

\begin{thebibliography}{49}
\providecommand{\natexlab}[1]{#1}
\providecommand{\url}[1]{\texttt{#1}}
\expandafter\ifx\csname urlstyle\endcsname\relax
  \providecommand{\doi}[1]{doi: #1}\else
  \providecommand{\doi}{doi: \begingroup \urlstyle{rm}\Url}\fi

\bibitem[Appelle and Lynch(2026)]{pedra2025}
Aaron Appelle and Jerome~P. Lynch.
\newblock {PEDRA}: Evaluating the realism of pedestrian dynamics in video
  generation.
\newblock In \emph{Proceedings of the IEEE/CVF Conference on Computer Vision
  and Pattern Recognition (CVPR)}, 2026.

\bibitem[Chen et~al.(2025)Chen, Zheng, Wang, Zhang, Zhan, Jia, Keutzer, and
  Zhang]{geodrive}
Anthony Chen, Wenzhao Zheng, Yida Wang, Xueyang Zhang, Kun Zhan, Peng Jia, Kurt
  Keutzer, and Shanghang Zhang.
\newblock {GeoDrive}: {3D} geometry-informed driving world model with precise
  action control.
\newblock \emph{arXiv preprint arXiv:2505.22421}, 2025.

\bibitem[Chen et~al.(2026)Chen, Zhang, Fu, Yang, Zhou, Zhang, He, Zhang, Sun,
  Dong, Long, Wang, and Meng]{drivingdino2026}
Xuyang Chen, Conglang Zhang, Chuanheng Fu, Zihao Yang, Kaixuan Zhou, Yizhi
  Zhang, Jianan He, Yanfeng Zhang, Mingwei Sun, Zhen Dong, Xiaoxiao Long,
  Zengmao Wang, and Liqiu Meng.
\newblock Driving with {DINO}: Vision foundation features as a unified bridge
  for sim-to-real generation in autonomous driving.
\newblock In \emph{Proceedings of the ACM International Conference on
  Multimedia}, 2026.

\bibitem[Cuevas-Velasquez et~al.(2024)Cuevas-Velasquez, Patel, Feng, and
  Black]{genbedlam2024}
Hanz Cuevas-Velasquez, Priyanka Patel, Haiwen Feng, and Michael~J. Black.
\newblock Toward human understanding with controllable synthesis.
\newblock \emph{arXiv preprint arXiv:2411.08663}, 2024.

\bibitem[Fang et~al.(2026)Fang, Xiang, Tan, Schuetz, Delp, Fei-Fei, and
  Adeli]{humanscore2026}
Yusu Fang, Tiange Xiang, Tian Tan, Narayan Schuetz, Scott Delp, Li Fei-Fei, and
  Ehsan Adeli.
\newblock {HumanScore}: Benchmarking human motions in generated videos.
\newblock \emph{arXiv preprint arXiv:2604.20157}, 2026.

\bibitem[Farley et~al.(2023)Farley, Zand, and Greenspan]{diffpeddet}
Andrew Farley, Mohsen Zand, and Michael~A. Greenspan.
\newblock Diffusion dataset generation: Towards closing the {Sim2Real} gap for
  pedestrian detection.
\newblock In \emph{2023 20th Conference on Robots and Vision (CRV)}, pages
  169--176, 2023.

\bibitem[Fu et~al.(2025)Fu, Hu, Zhou, Wang, Wang, and Liao]{pedvideoedit}
Danzhen Fu, Jiagao Hu, Daiguo Zhou, Fei Wang, Zepeng Wang, and Wenhua Liao.
\newblock Controllable pedestrian video editing for multi-view driving
  scenarios via motion sequence.
\newblock In \emph{Proceedings of the IEEE/CVF International Conference on
  Computer Vision Workshops (ICCVW)}, pages 1861--1870, 2025.

\bibitem[Gao et~al.(2026{\natexlab{a}})Gao, Yang, ang Gao, Li, Ding, Li, Yu,
  Liu, Xu, Niu, Chi, Chen, Tang, Zhang, Yi, and Zhao]{pam}
Mingju Gao, Kaisen Yang, Huan ang Gao, Bohan Li, Ao Ding, Wenyi Li, Yangcheng
  Yu, Jinkun Liu, Shaocong Xu, Yike Niu, Haohan Chi, Hao Chen, Hao Tang, Yu
  Zhang, Li Yi, and Hao Zhao.
\newblock {PAM}: A pose-appearance-motion engine for sim-to-real {HOI} video
  generation.
\newblock In \emph{Proceedings of the IEEE/CVF Conference on Computer Vision
  and Pattern Recognition (CVPR)}, pages 15954--15965, 2026{\natexlab{a}}.

\bibitem[Gao et~al.(2024{\natexlab{a}})Gao, Chen, Xie, Hong, Li, Yeung, and
  Xu]{magicdrive}
Ruiyuan Gao, Kai Chen, Enze Xie, Lanqing Hong, Zhenguo Li, Dit-Yan Yeung, and
  Qiang Xu.
\newblock {MagicDrive}: Street view generation with diverse {3D} geometry
  control.
\newblock In \emph{International Conference on Learning Representations
  (ICLR)}, 2024{\natexlab{a}}.

\bibitem[Gao et~al.(2025)Gao, Chen, Xiao, Hong, Li, and Xu]{magicdrivev2}
Ruiyuan Gao, Kai Chen, Bo Xiao, Lanqing Hong, Zhenguo Li, and Qiang Xu.
\newblock {MagicDrive-V2}: High-resolution long video generation for autonomous
  driving with adaptive control.
\newblock In \emph{Proceedings of the IEEE/CVF International Conference on
  Computer Vision (ICCV)}, pages 28135--28144, 2025.

\bibitem[Gao et~al.(2026{\natexlab{b}})Gao, Chen, Li, Hong, Li, and
  Xu]{magicdrive3d}
Ruiyuan Gao, Kai Chen, Zhihao Li, Lanqing Hong, Zhenguo Li, and Qiang Xu.
\newblock {MagicDrive3D}: Controllable {3D} generation for any-view rendering
  in street scenes.
\newblock In \emph{Proceedings of the IEEE/CVF Winter Conference on
  Applications of Computer Vision (WACV)}, pages 5724--5733,
  2026{\natexlab{b}}.

\bibitem[Gao et~al.(2024{\natexlab{b}})Gao, Yang, Chen, Chitta, Qiu, Geiger,
  Zhang, and Li]{vista}
Shenyuan Gao, Jiazhi Yang, Li Chen, Kashyap Chitta, Yihang Qiu, Andreas Geiger,
  Jun Zhang, and Hongyang Li.
\newblock {VISTA}: A generalizable driving world model with high fidelity and
  versatile controllability.
\newblock In \emph{Advances in Neural Information Processing Systems}, pages
  91560--91596, 2024{\natexlab{b}}.

\bibitem[Ge et~al.(2024)Ge, Mahapatra, Parmar, Zhu, and Huang]{cdfvd}
Songwei Ge, Aniruddha Mahapatra, Gaurav Parmar, Jun-Yan Zhu, and Jia-Bin Huang.
\newblock On the content bias in {Fr{\'e}chet} video distance.
\newblock In \emph{Proceedings of the IEEE/CVF Conference on Computer Vision
  and Pattern Recognition (CVPR)}, pages 7277--7288, 2024.

\bibitem[Geiger et~al.(2012)Geiger, Lenz, and Urtasun]{kitti}
Andreas Geiger, Philip Lenz, and Raquel Urtasun.
\newblock Are we ready for autonomous driving? the {KITTI} vision benchmark
  suite.
\newblock In \emph{Proceedings of the IEEE Conference on Computer Vision and
  Pattern Recognition (CVPR)}, pages 3354--3361, 2012.

\bibitem[Goel et~al.(2023{\natexlab{a}})Goel, Pavlakos, Rajasegaran, Kanazawa,
  and Malik]{fourdhumans}
Shubham Goel, Georgios Pavlakos, Jathushan Rajasegaran, Angjoo Kanazawa, and
  Jitendra Malik.
\newblock Humans in {4D}: Reconstructing and tracking humans with transformers.
\newblock In \emph{Proceedings of the IEEE/CVF International Conference on
  Computer Vision (ICCV)}, pages 14783--14794, 2023{\natexlab{a}}.

\bibitem[Goel et~al.(2023{\natexlab{b}})Goel, Pavlakos, Rajasegaran, Kanazawa,
  and Malik]{goel2023humans}
Shubham Goel, Georgios Pavlakos, Jathushan Rajasegaran, Angjoo Kanazawa, and
  Jitendra Malik.
\newblock Humans in {4D}: Reconstructing and tracking humans with transformers.
\newblock In \emph{Proceedings of IEEE/CVF International Conference on Computer
  Vision (ICCV)}, pages 14783--14794. IEEE, 2023{\natexlab{b}}.

\bibitem[G{\"u}ler et~al.(2018)G{\"u}ler, Neverova, and Kokkinos]{densepose}
R{\i}za~Alp G{\"u}ler, Natalia Neverova, and Iasonas Kokkinos.
\newblock {DensePose}: Dense human pose estimation in the wild.
\newblock In \emph{Proceedings of the IEEE Conference on Computer Vision and
  Pattern Recognition (CVPR)}, pages 7297--7306, 2018.

\bibitem[Guo et~al.(2025)Guo, Wu, Xiong, Xu, Zhou, Xu, Xu, Sun, Wang, Chen, Ye,
  Liu, and Wang]{genesis}
Xiangyu Guo, Zhanqian Wu, Kaixin Xiong, Ziyang Xu, Lijun Zhou, Gangwei Xu,
  Shaoqing Xu, Haiyang Sun, Bing Wang, Guang Chen, Hangjun Ye, Wenyu Liu, and
  Xinggang Wang.
\newblock {Genesis}: Multimodal driving scene generation with spatio-temporal
  and cross-modal consistency.
\newblock In \emph{Advances in Neural Information Processing Systems}, pages
  67252--67276, 2025.

\bibitem[Hassan et~al.(2025)Hassan, Stapf, Rahimi, Rezende, Haghighi,
  Br{\"u}ggemann, Katircioglu, Zhang, Chen, Saha, Cannici, Aljalbout, Ye, Wang,
  Davtyan, Salzmann, Scaramuzza, Pollefeys, Favaro, and Alahi]{gem}
Mariam Hassan, Sebastian Stapf, Ahmad Rahimi, Pedro M.~B. Rezende, Yasaman
  Haghighi, David Br{\"u}ggemann, Isinsu Katircioglu, Lin Zhang, Xiaoran Chen,
  Suman Saha, Marco Cannici, Elie Aljalbout, Botao Ye, Xi Wang, Aram Davtyan,
  Mathieu Salzmann, Davide Scaramuzza, Marc Pollefeys, Paolo Favaro, and
  Alexandre Alahi.
\newblock {GEM}: A generalizable ego-vision multimodal world model for
  fine-grained ego-motion, object dynamics, and scene composition control.
\newblock In \emph{Proceedings of the IEEE/CVF Conference on Computer Vision
  and Pattern Recognition (CVPR)}, pages 22404--22415, 2025.

\bibitem[He et~al.(2017)He, Gkioxari, Doll{\'a}r, and Girshick]{maskrcnn}
Kaiming He, Georgia Gkioxari, Piotr Doll{\'a}r, and Ross Girshick.
\newblock Mask {R-CNN}.
\newblock In \emph{Proceedings of the IEEE International Conference on Computer
  Vision (ICCV)}, pages 2961--2969, 2017.

\bibitem[Hu et~al.(2023)Hu, Russell, Yeo, Murez, Fedoseev, Kendall, Shotton,
  and Corrado]{gaia1}
Anthony Hu, Lloyd Russell, Hudson Yeo, Zak Murez, George Fedoseev, Alex
  Kendall, Jamie Shotton, and Gianluca Corrado.
\newblock {GAIA-1}: A generative world model for autonomous driving.
\newblock \emph{arXiv preprint arXiv:2309.17080}, 2023.

\bibitem[Hu(2024)]{animateanyone}
Li Hu.
\newblock {Animate Anyone}: Consistent and controllable image-to-video
  synthesis for character animation.
\newblock In \emph{Proceedings of the IEEE/CVF Conference on Computer Vision
  and Pattern Recognition (CVPR)}, pages 8153--8163, 2024.

\bibitem[Lee et~al.(2026)Lee, Lee, Kim, and Yoon]{veria}
Jumin Lee, Siyeong Lee, Namil Kim, and Sung-Eui Yoon.
\newblock {VERIA}: Verification-centric multimodal instance augmentation for
  long-tailed {3D} object detection.
\newblock \emph{arXiv preprint arXiv:2603.24294}, 2026.

\bibitem[Liang et~al.(2026{\natexlab{a}})Liang, Kong, Yan, Liu, Yang, Huang,
  Yin, Zuo, Hu, Zhu, Lu, Liu, Jiang, Li, Li, Zhuo, Ng, Cottereau, Gao, Pan,
  Ooi, and Liu]{worldlens}
Ao Liang, Lingdong Kong, Tianyi Yan, Hongsi Liu, Yu Yang, Ziqi Huang, Wei Yin,
  Jialong Zuo, Yixuan Hu, Dekai Zhu, Dongyue Lu, Youquan Liu, Guangfeng Jiang,
  Linfeng Li, Xiangtai Li, Long Zhuo, Lai~Xing Ng, Benoit~R. Cottereau,
  Changxin Gao, Liang Pan, Wei~Tsang Ooi, and Ziwei Liu.
\newblock {WorldLens}: Full-spectrum evaluations of driving world models in
  real world.
\newblock In \emph{Proceedings of the IEEE/CVF Conference on Computer Vision
  and Pattern Recognition (CVPR)}, pages 36385--36399, 2026{\natexlab{a}}.

\bibitem[Liang et~al.(2026{\natexlab{b}})Liang, Zhou, Li, Cao, Sun, Qian, Chen,
  and Wang]{realismotion}
Jingyun Liang, Jingkai Zhou, Shikai Li, Chenjie Cao, Lei Sun, Yichen Qian,
  Weihua Chen, and Fan Wang.
\newblock {RealisMotion}: Decomposed human motion control and video generation
  in the world space.
\newblock In \emph{Proceedings of the International Conference on Machine
  Learning (ICML)}, 2026{\natexlab{b}}.

\bibitem[Men et~al.(2025)Men, Yao, Cui, and Bo]{mimo}
Yifang Men, Yuan Yao, Miaomiao Cui, and Liefeng Bo.
\newblock {MIMO}: Controllable character video synthesis with spatial
  decomposed modeling.
\newblock In \emph{Proceedings of the IEEE/CVF Conference on Computer Vision
  and Pattern Recognition (CVPR)}, pages 21181--21191, 2025.

\bibitem[Mo et~al.(2025)Mo, Leng, Liu, Wang, He, and Zhou]{dreamland2025}
Sicheng Mo, Ziyang Leng, Leon Liu, Weizhen Wang, Honglin He, and Bolei Zhou.
\newblock {Dreamland}: Controllable world creation with simulator and
  generative models.
\newblock \emph{arXiv preprint arXiv:2506.08006}, 2025.

\bibitem[{NVIDIA}(2025{\natexlab{a}})]{cosmostransfer1}
{NVIDIA}.
\newblock {Cosmos-Transfer1}: Conditional world generation with adaptive
  multimodal control.
\newblock \emph{arXiv preprint arXiv:2503.14492}, 2025{\natexlab{a}}.

\bibitem[{NVIDIA}(2025{\natexlab{b}})]{cosmostransfer25}
{NVIDIA}.
\newblock World simulation with video foundation models for physical {AI}.
\newblock \emph{arXiv preprint arXiv:2511.00062}, 2025{\natexlab{b}}.

\bibitem[Ramesh et~al.(2026)Ramesh, Azer, and Flohr]{habit}
Mohan Ramesh, Mark Azer, and Fabian Flohr.
\newblock {HABIT}: Human action benchmark for interactive traffic in {CARLA}.
\newblock In \emph{Proceedings of the IEEE/CVF Winter Conference on
  Applications of Computer Vision (WACV)}, pages 7148--7157, 2026.

\bibitem[Ren et~al.(2025)Ren, Lu, Cao, Gao, Huang, Sabour, Shen, Pfaff, Wu,
  Chen, Kim, Gao, Leal-Taix{\'e}, Chen, Fidler, and Ling]{cosmosdrivedreams}
Xuanchi Ren, Yifan Lu, Tianshi Cao, Ruiyuan Gao, Shengyu Huang, Amirmojtaba
  Sabour, Tianchang Shen, Tobias Pfaff, Jay~Zhangjie Wu, Runjian Chen,
  Seung~Wook Kim, Jun Gao, Laura Leal-Taix{\'e}, Mike Chen, Sanja Fidler, and
  Huan Ling.
\newblock {Cosmos-Drive-Dreams}: Scalable synthetic driving data generation
  with world foundation models.
\newblock \emph{arXiv preprint arXiv:2506.09042}, 2025.

\bibitem[Russell et~al.(2025)Russell, Hu, Bertoni, Fedoseev, Shotton, Arani,
  and Corrado]{gaia2}
Lloyd Russell, Anthony Hu, Lorenzo Bertoni, George Fedoseev, Jamie Shotton,
  Elahe Arani, and Gianluca Corrado.
\newblock {GAIA-2}: A controllable multi-view generative world model for
  autonomous driving.
\newblock \emph{arXiv preprint arXiv:2503.20523}, 2025.

\bibitem[Shao et~al.(2024)Shao, Pang, Zheng, Sun, and Liu]{human4dit}
Ruizhi Shao, Youxin Pang, Zerong Zheng, Jingxiang Sun, and Yebin Liu.
\newblock {Human4DiT}: 360-degree human video generation with {4D} diffusion
  transformer.
\newblock \emph{ACM Transactions on Graphics}, 43\penalty0 (6):\penalty0 1--13,
  2024.

\bibitem[Wang et~al.(2024)Wang, Zhu, Huang, Chen, Zhu, and Lu]{drivedreamer}
Xiaofeng Wang, Zheng Zhu, Guan Huang, Xinze Chen, Jiagang Zhu, and Jiwen Lu.
\newblock {DriveDreamer}: Towards real-world-driven world models for autonomous
  driving.
\newblock In \emph{Proceedings of the European Conference on Computer Vision
  (ECCV)}, pages 55--72, 2024.

\bibitem[Wen et~al.(2024)Wen, Zhao, Liu, Jia, Wang, Luo, Zhang, Wang, Sun, and
  Zhang]{panacea}
Yuqing Wen, Yucheng Zhao, Yingfei Liu, Fan Jia, Yanhui Wang, Chong Luo, Chi
  Zhang, Tiancai Wang, Xiaoyan Sun, and Xiangyu Zhang.
\newblock {Panacea}: Panoramic and controllable video generation for autonomous
  driving.
\newblock In \emph{Proceedings of the IEEE/CVF Conference on Computer Vision
  and Pattern Recognition (CVPR)}, pages 6902--6912, 2024.

\bibitem[Wen et~al.(2026)Wen, Zhao, Liu, Huang, Jia, Wang, Zhang, Wang, Sun,
  and Zhang]{panaceaplus}
Yuqing Wen, Yucheng Zhao, Yingfei Liu, Binyuan Huang, Fan Jia, Yanhui Wang, Chi
  Zhang, Tiancai Wang, Xiaoyan Sun, and Xiangyu Zhang.
\newblock {Panacea+}: Panoramic and controllable video generation for
  autonomous driving.
\newblock \emph{IEEE Transactions on Circuits and Systems for Video
  Technology}, 36\penalty0 (2):\penalty0 2247--2258, 2026.

\bibitem[Wilson et~al.(2021)Wilson, Qi, Agarwal, Lambert, Singh, Khandelwal,
  Pan, Kumar, Hartnett, Pontes, Ramanan, Carr, and Hays]{argoverse2}
Benjamin Wilson, William Qi, Tanmay Agarwal, John Lambert, Jagjeet Singh,
  Siddhesh Khandelwal, Bowen Pan, Ratnesh Kumar, Andrew Hartnett,
  Jhony~Kaesemodel Pontes, Deva Ramanan, Peter Carr, and James Hays.
\newblock Argoverse 2: Next generation datasets for self-driving perception and
  forecasting.
\newblock In \emph{Proceedings of the Neural Information Processing Systems
  Track on Datasets and Benchmarks}, 2021.

\bibitem[Wu et~al.(2026)Wu, Pang, Chang, Ning, Wang, Xiao, and
  Li]{realitybridge2026}
Zhenhua Wu, Yun Pang, Mingkun Chang, Yuwei Ning, Liangzhi Wang, Yi Xiao, and
  Guanbin Li.
\newblock {RealityBridge}: Bridging editable {3D} gaussian splatting driving
  simulations and real-world videos.
\newblock \emph{arXiv preprint arXiv:2606.16278}, 2026.

\bibitem[Xu et~al.(2024)Xu, Zhang, Liew, Yan, Liu, Zhang, Feng, and
  Shou]{magicanimate}
Zhongcong Xu, Jianfeng Zhang, Jun~Hao Liew, Hanshu Yan, Jia-Wei Liu, Chenxu
  Zhang, Jiashi Feng, and Mike~Zheng Shou.
\newblock {MagicAnimate}: Temporally consistent human image animation using
  diffusion model.
\newblock In \emph{Proceedings of the IEEE/CVF Conference on Computer Vision
  and Pattern Recognition (CVPR)}, pages 1481--1490, 2024.

\bibitem[Yang and Zhang(2026)]{consisdrive}
Zhuoran Yang and Yanyong Zhang.
\newblock {ConsisDrive}: Identity-preserving driving world models for video
  generation by instance mask.
\newblock In \emph{International Conference on Learning Representations
  (ICLR)}, 2026.

\bibitem[Yang et~al.(2024)Yang, Guo, Ding, Wang, and Wu]{drivephysica}
Zhuoran Yang, Xi Guo, Chenjing Ding, Chiyu Wang, and Wei Wu.
\newblock Physical informed driving world model.
\newblock \emph{arXiv preprint arXiv:2412.08410}, 2024.

\bibitem[Zhang et~al.(2025)Zhang, Gu, Wang, Wang, Cheng, Zhu, and
  Zou]{mimicmotion}
Yuang Zhang, Jiaxi Gu, Li-Wen Wang, Han Wang, Junqi Cheng, Yuefeng Zhu, and
  Fangyuan Zou.
\newblock {MimicMotion}: High-quality human motion video generation with
  confidence-aware pose guidance.
\newblock In \emph{Proceedings of the International Conference on Machine
  Learning (ICML)}, pages 74896--74910, 2025.

\bibitem[Zhao et~al.(2025)Zhao, Wang, Zhu, Chen, Huang, Bao, and
  Wang]{drivedreamer2}
Guosheng Zhao, Xiaofeng Wang, Zheng Zhu, Xinze Chen, Guan Huang, Xiaoyi Bao,
  and Xingang Wang.
\newblock {DriveDreamer-2}: {LLM}-enhanced world models for diverse driving
  video generation.
\newblock In \emph{Proceedings of the AAAI Conference on Artificial
  Intelligence}, pages 10412--10420, 2025.

\bibitem[Zhao et~al.(2026)Zhao, Wang, Chen, Donzella, Bashford-Rogers, and
  Debattista]{drivectrl2026}
Haonan Zhao, Yiting Wang, Jingkun Chen, Valentina Donzella, Thomas
  Bashford-Rogers, and Kurt Debattista.
\newblock {DriveCtrl}: Conditioned sim-to-real driving video generation.
\newblock \emph{arXiv preprint arXiv:2605.15116}, 2026.

\bibitem[Zheng et~al.(2026)Zheng, Li, Deng, Wang, Chen, Xiao, Chi, Lin, Chen,
  Wang, Zhang, and Liu]{xworld}
Chaoda Zheng, Sean Li, Jinhao Deng, Zhennan Wang, Shijia Chen, Liqiang Xiao,
  Ziheng Chi, Hongbin Lin, Kangjie Chen, Boyang Wang, Yu Zhang, and Xianming
  Liu.
\newblock {X-World}: Controllable ego-centric multi-camera world models for
  scalable end-to-end driving.
\newblock \emph{arXiv preprint arXiv:2603.19979}, 2026.

\bibitem[Zhou et~al.(2025)Zhou, Wu, Li, Wei, Fan, Chen, Jiang, and
  Wang]{realisdancedit2025}
Jingkai Zhou, Yifan Wu, Shikai Li, Min Wei, Chao Fan, Weihua Chen, Wei Jiang,
  and Fan Wang.
\newblock {RealisDance-DiT}: Simple yet strong baseline towards controllable
  character animation in the wild.
\newblock \emph{arXiv preprint arXiv:2504.14977}, 2025.

\bibitem[Zhou et~al.(2024)Zhou, Simon, Peng, Mo, Zhu, Guo, and Zhou]{simgen}
Yunsong Zhou, Michael Simon, Zhenghao Peng, Sicheng Mo, Hongzi Zhu, Minyi Guo,
  and Bolei Zhou.
\newblock {SimGen}: Simulator-conditioned driving scene generation.
\newblock In \emph{Advances in Neural Information Processing Systems}, 2024.

\bibitem[Zhu et~al.(2025)Zhu, Fan, Dai, and Ho]{waymo3dskelmo}
Guangxun Zhu, Shiyu Fan, Hang Dai, and Edmond S.~L. Ho.
\newblock {Waymo-3DSkelMo}: A multi-agent {3D} skeletal motion dataset for
  pedestrian interaction modeling in autonomous driving.
\newblock In \emph{Proceedings of the ACM International Conference on
  Multimedia}, pages 13184--13190, 2025.

\bibitem[Zhu et~al.(2024)Zhu, Chen, Dai, Dong, Xu, Cao, Yao, Zhu, and
  Zhu]{champ}
Shenhao Zhu, Junming~Leo Chen, Zuozhuo Dai, Zilong Dong, Yinghui Xu, Xun Cao,
  Yao Yao, Hao Zhu, and Siyu Zhu.
\newblock {Champ}: Controllable and consistent human image animation with {3D}
  parametric guidance.
\newblock In \emph{Proceedings of the European Conference on Computer Vision
  (ECCV)}, pages 145--162, 2024.

\end{thebibliography}
}

\clearpage
\setcounter{section}{0}
\setcounter{subsection}{0}
\setcounter{table}{0}
\setcounter{figure}{0}
\setcounter{equation}{0}
\renewcommand{\thesection}{S\arabic{section}}
\renewcommand{\thetable}{S\arabic{table}}
\renewcommand{\thefigure}{S\arabic{figure}}
\renewcommand{\theequation}{S\arabic{equation}}
\renewcommand{\theHsection}{supp.\arabic{section}}
\renewcommand{\theHtable}{supp.\arabic{table}}
\renewcommand{\theHfigure}{supp.\arabic{figure}}
\renewcommand{\theHequation}{supp.\arabic{equation}}
\section*{Supplementary Material}

\section{Simulator-Grounded Identity Diagnostics}
\label{sec:supp-evaluation}

The main paper defines the simulator-grounded pose metrics and
top-down evaluation protocol. Here, we specify the additional
diagnostics used to characterize identity-related failures:
the mutually exclusive failure taxonomy of Main Paper Fig.~4 and the
separate facing-direction test of Main Paper Fig.~5.
\begin{figure*}[ht]
    \centering
    \includegraphics[width=\textwidth]{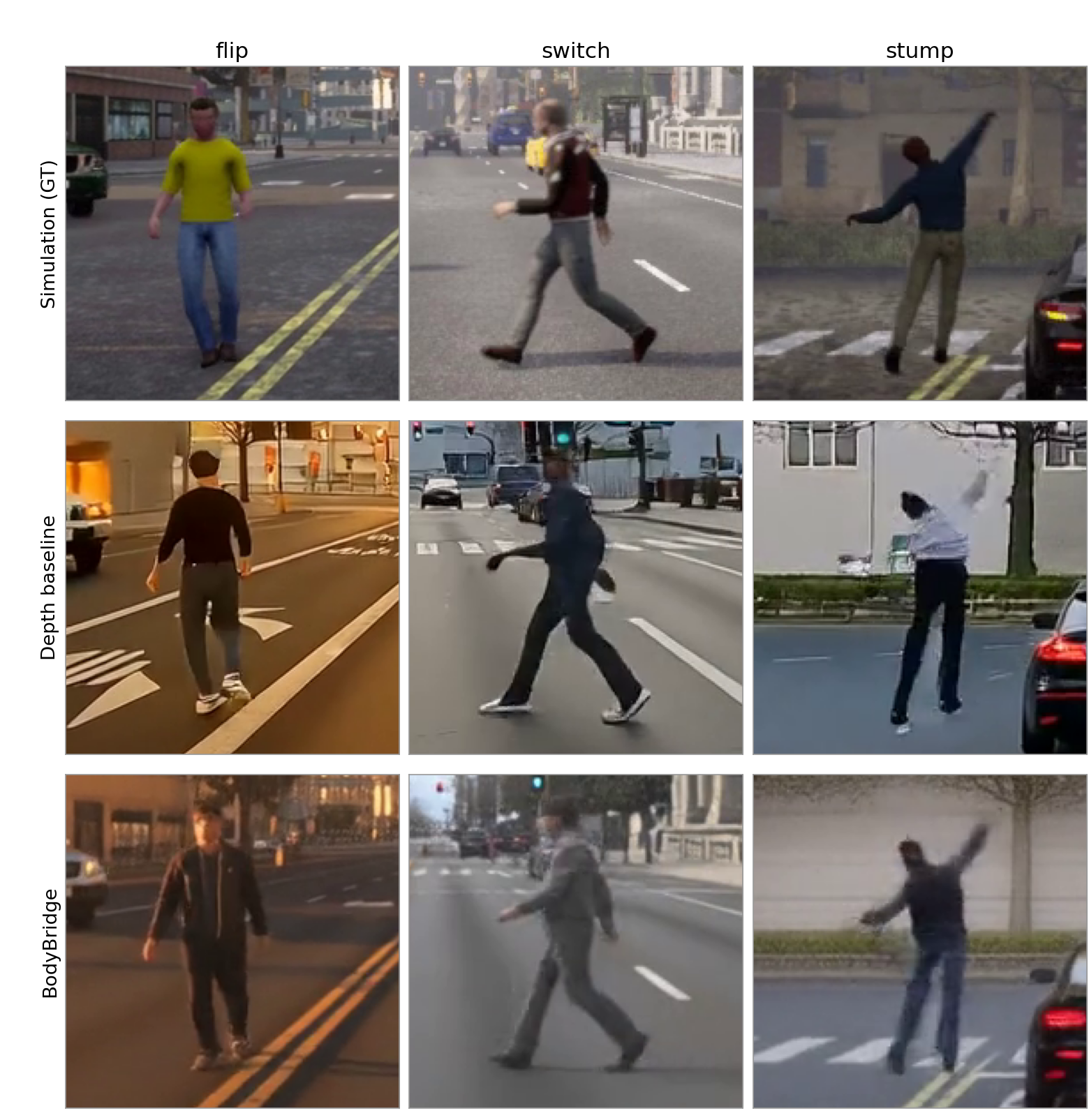}
    \caption{\textbf{Matched qualitative examples of taxonomy-defined
    articulation failures.}
    Columns show representative \emph{flip}, \emph{switch}, and
    \emph{stump} cases. The top row shows the corresponding simulator
    ground truth, the middle row the released Cosmos depth-control
    baseline, and the bottom row BodyBridge. The examples illustrate
    how the taxonomy manifests visually under matched source states.
    Reported failure frequencies are computed over the full evaluation set rather than from these examples.}
    \label{fig:supp-taxonomy-qual}
\end{figure*}
\subsection{Failure-Taxonomy Definitions}
\label{sec:supp-taxonomy}

The taxonomy is applied to GT-valid COCO-17 keypoints
that fail the PCK@0.5 criterion. For instance $n$ and joint $j$,
let $\mathbf g_{n,j}$ and $\hat{\mathbf g}_{n,j}$ denote the projected
simulator-GT and predicted COCO-17 keypoints, respectively, following
Main Paper Sec.~3.4. We define the identity-preserving localization error
\begin{equation}
d^{\mathrm{id}}_{n,j}
=
\left\|
\hat{\mathbf g}_{n,j}-\mathbf g_{n,j}
\right\|_2 ,
\label{eq:supp-id-error}
\end{equation}
and let $\sigma(j)$ denote the corresponding left--right COCO label.
The error after exchanging the predicted left--right identity is
\begin{equation}
d^{\mathrm{swap}}_{n,j}
=
\left\|
\hat{\mathbf g}_{n,\sigma(j)}
-
\mathbf g_{n,j}
\right\|_2 .
\label{eq:supp-swap-error}
\end{equation}
All thresholds are normalized by the GT torso diagonal $d_n$ from
Main Paper Eq.~(8). At PCK@0.5, the localization threshold is
$\tau_n=0.5d_n$. A GT-valid joint is considered failed if its
prediction is unavailable or non-finite, or if
$d^{\mathrm{id}}_{n,j}>\tau_n$.

Each failed joint receives exactly one label according to the fixed
priority
\begin{equation}
\mathrm{miss}
>
\mathrm{flip}
>
\mathrm{switch}
>
\mathrm{merge}
>
\mathrm{stump}
>
\mathrm{residual}.
\label{eq:supp-priority}
\end{equation}
Thus, when multiple conditions apply to the same failed joint, only the
highest-priority category is assigned.

A \emph{miss} is assigned when the corresponding prediction is
unavailable or non-finite.

A \emph{full-body flip} is detected at the instance level. The test
uses the left--right shoulder, elbow, wrist, hip, knee, and ankle
pairs for which the required GT and predicted joints are available.
At least four eligible joints spanning at least two distinct body
pairs are required. Let $\mathcal P_n$ denote these eligible joints.
The instance is classified as flipped when
\begin{equation}
\begin{aligned}
\operatorname*{med}_{j\in\mathcal P_n}
d^{\mathrm{swap}}_{n,j}
&<
0.75\,
\operatorname*{med}_{j\in\mathcal P_n}
d^{\mathrm{id}}_{n,j},
\\
\operatorname*{med}_{j\in\mathcal P_n}
d^{\mathrm{id}}_{n,j}
&>
\tau_n .
\end{aligned}
\label{eq:supp-flip}
\end{equation}
Failed joints on such an instance receive the flip label before any
pair- or limb-level condition is considered. Sweeping the ratio
threshold in Eq.~\eqref{eq:supp-flip} from $0.60$ to $0.90$
preserves the reported reduction trend.

A \emph{switch} is evaluated only when the instance is not classified
as a full-body flip. For a left--right pair $(l,r)$, both joints must
be GT-valid and detected. A switch requires the identity-preserving
assignment to fail for at least one joint,
\begin{equation}
d^{\mathrm{id}}_{n,l}>\tau_n
\quad\text{or}\quad
d^{\mathrm{id}}_{n,r}>\tau_n ,
\label{eq:supp-switch-fail}
\end{equation}
while exchanging only that pair's predicted identities places both
joints within threshold:
\begin{equation}
\begin{aligned}
\left\|
\hat{\mathbf g}_{n,r}-\mathbf g_{n,l}
\right\|_2
&\leq \tau_n,
\\
\left\|
\hat{\mathbf g}_{n,l}-\mathbf g_{n,r}
\right\|_2
&\leq \tau_n .
\end{aligned}
\label{eq:supp-switch}
\end{equation}
A switch therefore denotes a pair that is incorrect as labeled but
becomes correct after exchanging that pair's left--right identities.

A \emph{merge} measures collapse of a detected left--right pair.
For an eligible pair, define its GT and predicted separations as
\begin{equation}
\begin{aligned}
s_n^{\mathrm{GT}}
&=
\left\|
\mathbf g_{n,l}-\mathbf g_{n,r}
\right\|_2,
\\
s_n^{\mathrm{pred}}
&=
\left\|
\hat{\mathbf g}_{n,l}-\hat{\mathbf g}_{n,r}
\right\|_2 .
\end{aligned}
\label{eq:supp-separation}
\end{equation}
Both joints must be GT-valid and detected, and the GT separation must
satisfy
\begin{equation}
s_n^{\mathrm{GT}}>0.15d_n .
\end{equation}
A merge is detected when
\begin{equation}
s_n^{\mathrm{pred}}
<
0.30\,s_n^{\mathrm{GT}} .
\label{eq:supp-merge}
\end{equation}
The merge condition is evaluated independently of the instance-level
flip flag, but the attribution priority in
Eq.~\eqref{eq:supp-priority} ensures that failed joints on a flipped
instance remain labeled as flips.

A \emph{stump} is evaluated on the four three-joint limb chains
\[
(\mathrm{LShoulder,LElbow,LWrist}),\;
(\mathrm{RShoulder,RElbow,RWrist}),
\]
\[
(\mathrm{LHip,LKnee,LAnkle}),\;
(\mathrm{RHip,RKnee,RAnkle}).
\]
For chain $(a,b,c)$, all three joints must be GT-valid and detected.
Its two-segment GT length is
\begin{equation}
\ell_n^{\mathrm{GT}}
=
\left\|
\mathbf g_{n,b}-\mathbf g_{n,a}
\right\|_2
+
\left\|
\mathbf g_{n,c}-\mathbf g_{n,b}
\right\|_2 ,
\label{eq:supp-chain-length}
\end{equation}
with $\ell_n^{\mathrm{pred}}$ defined analogously from the predictions.
The chain is eligible when
\begin{equation}
\ell_n^{\mathrm{GT}}>0.20d_n ,
\end{equation}
and is classified as a stump when
\begin{equation}
\ell_n^{\mathrm{pred}}
<
0.40\,\ell_n^{\mathrm{GT}} .
\label{eq:supp-stump}
\end{equation}
Only failed mid and distal joints---elbow/wrist or knee/ankle---receive
the stump label; the proximal shoulder or hip does not.

Any failed joint that satisfies none of the preceding conditions is
assigned to the \emph{residual} category. The merge and stump
thresholds are fixed protocol values rather than parameters selected
from the evaluation results.  

Figure~\ref{fig:supp-taxonomy-qual} shows matched qualitative examples
of three taxonomy-defined failure types---flip, switch, and stump---for
the released depth baseline and BodyBridge, using the corresponding
simulator frame as reference.

\subsection{Facing-Direction Protocol}
\label{sec:supp-facing}

Facing direction is evaluated separately from the failure taxonomy
because it tests front--back orientation rather than keypoint
localization. For a recovered instance $n$, we use the horizontal
left--right shoulder ordering. Define
\begin{equation}
\begin{aligned}
\Delta_n^{\mathrm{GT}}
&=
g_{n,\mathrm{LShoulder},x}
-
g_{n,\mathrm{RShoulder},x},
\\
\Delta_n^{\mathrm{pred}}
&=
\hat g_{n,\mathrm{LShoulder},x}
-
\hat g_{n,\mathrm{RShoulder},x}.
\end{aligned}
\label{eq:supp-facing-delta}
\end{equation}

An instance is considered \emph{facing-checkable} only when both
shoulders are available and their projected separation is sufficiently
large in both the GT and prediction:
\begin{equation}
\begin{aligned}
\left|\Delta_n^{\mathrm{GT}}\right|
&>0.10d_n,
\\
\left|\Delta_n^{\mathrm{pred}}\right|
&>0.10d_n .
\end{aligned}
\label{eq:supp-facing-gate}
\end{equation}
This excludes near-profile configurations for which shoulder ordering
does not provide a reliable front--back cue.

For the remaining instances, the facing-flip indicator is
\begin{equation}
f_n
=
\mathds{1}
\left[
\operatorname{sgn}
\!\left(\Delta_n^{\mathrm{GT}}\right)
\neq
\operatorname{sgn}
\!\left(\Delta_n^{\mathrm{pred}}\right)
\right].
\label{eq:supp-facing}
\end{equation}
The diagnostic therefore tests whether the generated VRU preserves the
simulator-specified front--back orientation without requiring agreement
in absolute shoulder position.

Instances that fail the gate in Eq.~\eqref{eq:supp-facing-gate}
receive no facing label rather than being counted as correct or
incorrect. Facing-flip rates are consequently computed only over
recovered, facing-checkable instances. The same frame-wise indicator is
used for the temporal analysis in Supplementary Sec.~S3.1.

\section{Robustness and Conditioning Audit}
\label{sec:supp-robustness}

The main paper reports the primary simulator-grounded comparisons.
Here, we test whether their interpretation changes under inference
stochasticity, independently trained control branches, stricter pose
thresholds, a different pose estimator, or error in the
body-registration procedure. These analyses vary one source of
uncertainty at a time while otherwise retaining the evaluation
protocol of the main paper.

\subsection{Inference-Seed Replication}
\label{sec:supp-inference-seeds}

To isolate stochasticity during video generation, we regenerate the
five principal configurations with inference seeds 42, 43, and 44,
while keeping the trained checkpoints, conditioning inputs, prompt,
and inference settings fixed. Table~\ref{tab:supp-inference-seeds}
reports near-range PCK@0.5 for each generated sample.

\begin{table}[t]
\centering
\caption{\textbf{Inference-seed replication.}
Near-range PCK@0.5 for three inference seeds with trained checkpoints
held fixed.}
\label{tab:supp-inference-seeds}
\small
\begin{tabular}{lcccc}
\toprule
Configuration & s42 & s43 & s44 & Mean \\
\midrule
Cosmos depth              & .666 & .696 & .655 & .673 \\
Normal map                & .796 & .793 & .789 & .793 \\
Canonical corr.  & .805 & .802 & .795 & .800 \\
Depth + normal map        & .808 & .819 & .791 & .806 \\
BodyBridge                & .812 & .828 & .803 & .814 \\
\bottomrule
\end{tabular}
\end{table}

BodyBridge remains above the released depth-control baseline for all
three inference seeds. The corresponding per-seed near-PCK differences
are $+0.146$, $+0.132$, and $+0.148$. The canonical-correspondence and
normal-map branches, in contrast, remain close in aggregate near-range
PCK across all three samples. This separation is consistent with the
main-paper interpretation that BodyBridge's advantage over scene
geometry is robust at the aggregate pose level, whereas the distinction
between the two dense body representations is primarily
identity-specific.

\subsection{Training-Run Replication}
\label{sec:supp-training-runs}

Inference-seed replication leaves the trained control branch fixed.
To assess optimization variability separately, we evaluate three
independently trained canonical-correspondence branches produced with
the same architecture and training recipe. The main-paper checkpoint
was fixed before the two replication runs were trained; no checkpoint
selection was performed across the three runs.

\begin{table}[t]
\centering
\caption{\textbf{Training-run replication.}
Near-range PCK@0.5 and relative identity-error reductions for three
independently trained canonical-correspondence branches. The reduction
columns compare canonical correspondence with normal-map conditioning
under the same evaluation protocol.}
\label{tab:supp-training-runs}
\small
\begin{tabular}{lccc}
\toprule
Training run &
Near PCK $\uparrow$ &
Flip red. $\uparrow$ &
Facing red. $\uparrow$ \\
\midrule
Original  & .805 & 18.4\% & 21.6\% \\
Replica 1 & .771 & 24.2\% & 17.4\% \\
Replica 2 & .779 & 20.8\% & 13.0\% \\
\bottomrule
\end{tabular}
\end{table}

Near-range PCK varies from $0.771$ to $0.805$, showing that aggregate
pose fidelity is sensitive to control-branch training. The
identity-specific effect is more consistent: full-body-flip reduction
remains between $18.4\%$ and $24.2\%$, and facing-flip reduction
remains between $13.0\%$ and $21.6\%$, with the same direction in all
three runs. The replication therefore supports the narrower claim made
in the main paper: canonical correspondence consistently reduces the
targeted identity errors, while its aggregate PCK difference from the
normal-map representation is comparatively small and training-run
dependent.

\subsection{Recovery and Distance-Stratified Results}
\label{sec:supp-recovery-distance}

Pose fidelity and instance recovery capture different failure modes.
Table~\ref{tab:supp-recovery} therefore reports recovery separately
across the simulator-distance strata for the five principal
configurations.

\begin{table}[t]
\centering
\caption{\textbf{Distance-stratified instance recovery.}
Fraction of scorable GT VRUs recovered under the top-down evaluation
protocol.}
\label{tab:supp-recovery}
\small
\begin{tabular}{lcccc}
\toprule
Configuration &
All $\uparrow$ &
Near $\uparrow$ &
Mid $\uparrow$ &
Far $\uparrow$ \\
\midrule
Cosmos depth              & .504 & .906 & .640 & .229 \\
Normal map                & .414 & .943 & .561 & .082 \\
Canonical corr.           & .431 & .938 & .575 & .109 \\
Depth + normal map        & .486 & .955 & .663 & .146 \\
BodyBridge                & .493 & .943 & .677 & .167 \\
\bottomrule
\end{tabular}
\end{table}

The body-only branches retain high near-range recovery but lose
coverage rapidly with distance. Scene-geometry composition changes
this behavior. For canonical correspondence, composition raises
mid-range recovery from $0.575$ to $0.677$ and far-range recovery from
$0.109$ to $0.167$, while near-range recovery remains similar
($0.938$ versus $0.943$). The corresponding normal-map composition
shows the same coverage effect.

The distance breakdown therefore clarifies the role of composition:
body conditioning provides the articulated signal, whereas
scene-geometry control primarily restores recovery as projected VRUs
become smaller. This is why recovery is kept separate from the
pose-fidelity measures rather than interpreting changes in pooled PCK
as articulation alone.

\subsection{PCK-Threshold Sensitivity}
\label{sec:supp-pck-threshold}

We test whether the primary BodyBridge--depth ordering depends on the
PCK@0.5 localization tolerance by repeating the paired near-range
comparison at substantially stricter thresholds.

\begin{table}[t]
\centering
\caption{\textbf{PCK-threshold sensitivity.}
Paired near-range PCK difference between BodyBridge and the released
depth-control baseline. Positive values favor BodyBridge.}
\label{tab:supp-pck-threshold}
\small
\begin{tabular}{ccc}
\toprule
Threshold & $\Delta$PCK & 95\% CI \\
\midrule
$\alpha=.5$ & +.146 & [+.097, +.195] \\
$\alpha=.2$ & +.148 & [+.101, +.187] \\
$\alpha=.1$ & +.069 & [+.045, +.088] \\
\bottomrule
\end{tabular}
\end{table}

The difference remains positive at all three thresholds and is similar
at $\alpha=.5$ and $\alpha=.2$. It also remains resolved when the
tolerance is reduced to $\alpha=.1$. The BodyBridge--depth ordering is
therefore not specific to the PCK@0.5 threshold used for the primary
evaluation.

\subsection{Recovery Uncertainty}
\label{sec:supp-recovery-uncertainty}

The distance-stratified values in Table~\ref{tab:supp-recovery} are
interpreted primarily as coverage diagnostics. Statistical inference
on recovery is restricted to the near-range comparison used alongside
the primary pose analysis; its paired confidence interval is reported
in the main paper. Pooled, mid-, and far-range recovery are retained
as descriptive measures because their main role is to expose how
coverage changes with distance rather than to support separate
performance claims.

This distinction is particularly important at far range, where
recovery is strongly affected by the spatial resolution available to
the pose estimator. We therefore do not use the far-range recovery
difference as evidence for or against articulated-body fidelity.

\subsection{Cross-Estimator Robustness}
\label{sec:supp-cross-estimator}

The main evaluation uses YOLO11-L-pose. To test whether the
simulator-grounded ordering depends on this particular estimator, we
repeat the complete top-down evaluation with Keypoint R-CNN~\citep{maskrcnn}
(R50-FPN), using the same GT boxes, projected simulator keypoints,
distance strata, and scoring rules.

Across generation configurations, the alternative estimator preserves
the ordering of the metrics most relevant to the main claim. Spearman
agreement between the two estimators is $\rho=0.98$ for recovery,
$\rho=0.95$ for near-range PCK@0.5, $\rho=0.93$ for NMJE, and
$\rho=0.91$ for far-range PCK. Agreement is weaker for pooled PCK
($\rho=0.52$) and mid-range PCK ($\rho=0.60$).

Under Keypoint R-CNN, the paired BodyBridge--depth near-range
difference remains positive:
\begin{equation}
\begin{aligned}
\Delta\mathrm{PCK}_{\mathrm{near}} &= +0.090,\\
95\%~\mathrm{CI} &= [+0.020,+0.150].
\end{aligned}
\label{eq:supp-cross-estimator}
\end{equation}
The corresponding single-inference-seed difference under
YOLO11-L-pose is $+0.146$. Thus, the magnitude changes with the
estimator, but the primary near-range ordering does not.

Pooled PCK is less estimator-stable and changes sign under
Keypoint R-CNN. We therefore retain it as a coverage-weighted summary
rather than using it as evidence for the primary BodyBridge--depth
comparison. The cross-estimator result instead supports the near-range
pose comparison on which the main claim is based.

\subsection{Registration and Conditioning-Geometry}
\label{sec:supp-registration}

Canonical correspondence requires aligning the posed body template to
the corresponding simulator actor before rasterization. This alignment
affects the conditioning signal but not the simulator-derived
evaluation reference. Registration error can therefore degrade the
control signal, but cannot improve the GT against which the generated
video is scored.

For each actor--frame pair with the required body-pose representation,
we render the registered body in isolation and compare its visible
silhouette with the simulator-provided actor mask. Visibility is
restricted using simulator depth. We additionally measure the
image-space reprojection residual of the registered body joints.

\begin{table}[t]
\centering
\caption{\textbf{Registration and conditioning-geometry audit.}
Visible-mask IoU compares the registered body render with the simulator
actor mask. Joint reprojection RMS is reported in pixels and relative
to apparent body height.}
\label{tab:supp-registration}
\small
\setlength{\tabcolsep}{4.5pt}
\begin{tabular}{lccc}
\toprule
Distance &
IoU $\uparrow$ &
RMS [px] $\downarrow$ &
RMS [\% height] $\downarrow$ \\
\midrule
Near & .685 [.652,.710] & 3.1 & 3.7 \\
Mid  & .645 [.628,.659] & 1.5 & 3.8 \\
Far  & .581 [.564,.597] & 0.8 & 3.3 \\
\bottomrule
\end{tabular}
\end{table}

Absolute reprojection residual decreases with distance as the apparent
body size becomes smaller, while the scale-normalized error remains
stable at approximately $3.3$--$3.8\%$ of body height. Visible-mask
IoU is highest in the near range and decreases as the projected actors
become smaller.

We also test whether the frame-wise similarity fit introduces temporal
variation. For the fitted similarity scale $s$, the median relative frame-to-frame change in fitted scale is
approximately
\[
\operatorname{median}
\frac{|\Delta s|}{s}
\approx 0.1\%.
\]
For near-range tracks, the median pelvis-motion residual attributable
to registration is $0.11$ pixels, compared with $4.3$ pixels of actual
frame-to-frame pelvis motion.

The required body-pose representation is available for $97.1\%$ of
conditioning actor--frame pairs. The remaining $2.9\%$ use the
2D-skeleton fallback described in the main paper. These cases arise
from missing pose-parameter coverage rather than failure of the
similarity registration: among cases for which the required
representation is available, no fit fails because of insufficient
correspondences or a singular Umeyama solution. The fallback changes
only the conditioning representation; the simulator-derived evaluation
reference remains unchanged.

\section{Temporal Analysis}
\label{sec:supp-temporal}

The frame-wise metrics in the main paper evaluate whether each
generated VRU preserves the simulator-specified articulation at a
given time. Here, we complement them with a trajectory-level analysis
that tests whether this agreement is also preserved over time.

\subsection{Temporal Pose and Identity Consistency}
\label{sec:supp-temporal-consistency}

We evaluate temporal articulation using acceleration error on keypoint
trajectories that are jointly scorable across compared generation
variants. To separate articulated motion from global image-plane
translation, the instance centroid is removed independently at each
frame. For GT and predicted centroid-relative keypoint positions
$\mathbf g^{\mathrm{rel}}_{t,j}$ and
$\hat{\mathbf g}^{\mathrm{rel}}_{t,j}$, respectively, define
\begin{equation}
\begin{aligned}
\Delta^2 \mathbf g^{\mathrm{rel}}_{t,j}
&=
\mathbf g^{\mathrm{rel}}_{t+1,j}
-
2\mathbf g^{\mathrm{rel}}_{t,j}
\\
&\quad+
\mathbf g^{\mathrm{rel}}_{t-1,j}.
\end{aligned}
\label{eq:supp-second-difference}
\end{equation}
The normalized acceleration error is then
\begin{equation}
E^{\mathrm{rel}}_{\mathrm{acc}}
=
\frac{
\left\|
\Delta^2\hat{\mathbf g}^{\mathrm{rel}}_{t,j}
-
\Delta^2\mathbf g^{\mathrm{rel}}_{t,j}
\right\|_2
}{
d_t
},
\label{eq:supp-acceleration}
\end{equation}
where $d_t$ is the GT torso normalization scale. The metric is
evaluated only on strictly consecutive frame triples with matched
temporal support across the compared variants.

Relative to the released depth-control baseline, BodyBridge reduces
centroid-relative acceleration error by
\begin{equation}
\begin{aligned}
\Delta E^{\mathrm{rel}}_{\mathrm{acc}} &= -0.0633,\\
95\%~\mathrm{CI} &= [-0.1041,-0.0267].
\end{aligned}
\label{eq:supp-acceleration-result}
\end{equation}
The reduction is also present in the near range
($\Delta E^{\mathrm{rel}}_{\mathrm{acc}}=-0.0334$).
The body-only canonical-correspondence branch shows the same direction
before composition, with an overall difference of $-0.0345$
(95\% CI $[-0.0616,-0.0083]$) and a near-range difference of
$-0.0326$. These paired comparisons are computed per track with
clip-clustered bootstrap resampling.

We separately examine temporal behavior of the facing-direction
indicator defined in Supplementary Sec.~S1.2. On the matched set of
$390$ near-range tracks, the frame-wise facing-flip rate is $0.162$
for BodyBridge and $0.530$ for depth. Among admissible adjacent scored
frames, the fraction for which the binary facing-error state changes is
$0.069$ for BodyBridge and $0.128$ for depth.

The latter quantity is a descriptive identity-stability diagnostic,
not a standard temporal-pose metric. Together with acceleration error,
it distinguishes two aspects of temporal behavior: acceleration error
measures agreement with the simulator-specified articulated trajectory,
whereas the facing diagnostic tests whether an incorrect front--back
identity persists or changes across frames.

\section{Application and Real-Data Controls}
\label{sec:supp-application}

The main paper reports the downstream detection association and the
real-video extension. Here, we expose the detector measurements behind
the downstream analysis, test whether the association depends on the
untranslated simulator reference, and document the quality-control and
data-separation procedures used for pseudo-labelled real-video
supervision.

\subsection{Downstream Detection Results}
\label{sec:supp-downstream-results}

Table~\ref{tab:supp-downstream-results} reports downstream detector
performance for five principal image-based generation variants.
For generation configuration $m$ and detector seed $s$, the detector
gain is
\begin{equation}
\Delta\mathrm{AP}_{m,s}
=
\mathrm{AP}_{m,s}
-
\mathrm{AP}_{\mathrm{real},s},
\label{eq:supp-delta-ap}
\end{equation}
where the reference is the real-only detector trained with the same
seed. The reported gains are means over six detector seeds. All
conditions use the same augmentation frame indices and
simulator-derived GT boxes and are evaluated on the same held-out real
KITTI validation split. Rows are ordered by
$\Delta$AP@[.5:.95], from the smallest to the largest performance loss.

\begin{table}[t]
\centering
\caption{\textbf{Pose fidelity and downstream detection.}
Near-range PCK@0.5 is measured on each generated-data variant.
Detector gains are means over six YOLO11-s training seeds relative
to the corresponding same-seed real-only baseline. All variants use
identical augmentation frames and GT boxes. Rows are ordered by
$\Delta$AP@[.5:.95].}
\label{tab:supp-downstream-results}
\small
\setlength{\tabcolsep}{3.0pt}
\begin{tabular}{lccc}
\toprule
Configuration &
Near PCK $\uparrow$ &
$\Delta$AP@.5 $\uparrow$ &
$\Delta$AP@[.5:.95] $\uparrow$ \\
\midrule
BodyBridge      & .812 & $-.0095$ & $-.0051$ \\
Depth           & .666 & $-.0137$ & $-.0068$ \\
Normal map      & .738 & $-.0113$ & $-.0077$ \\
Edge            & .492 & $-.0233$ & $-.0114$ \\
2D skeleton     & .595 & $-.0261$ & $-.0139$ \\
\bottomrule
\end{tabular}
\end{table}

Among the displayed variants, BodyBridge has the
smallest AP@[.5:.95] deficit, whereas edge and 2D-skeleton generation
produce the largest losses. The ordering is not strictly determined by
near-range PCK alone: depth has lower near-range PCK than the normal-map
variant but a smaller AP@[.5:.95] deficit. This is expected because
generated appearance and other visual properties can also affect
detector training.

The rank analysis in Main Paper Sec.~5.4 is computed over the complete
predefined configuration set; Table~\ref{tab:supp-downstream-results}
shows the principal image-based variants rather than redefining that
analysis. Since augmentation quantity, frame selection, and GT boxes
are held fixed, those factors cannot explain the detector differences.
Other visual properties may still co-vary with pose fidelity, so the
experiment is interpreted as an association between simulator-grounded
pose fidelity and detector utility rather than as a causal estimate of
the effect of pose error.

As a robustness check, we repeat the rank analysis after excluding the
untranslated sim-RGB reference, whose exact simulator articulation is
paired with a large appearance-domain gap to KITTI. Across the remaining
eight configurations, the association between near PCK@0.5 and mean
$\Delta$AP@0.5 remains positive
($\rho=0.821$, exact permutation $p=0.034$).
Thus, the downstream association is not driven solely by the
high-pose-fidelity simulator endpoint. Its stability across detector
training seeds is reported in Main Paper Sec.~5.4.

\subsection{Pseudo-Label and Appearance-Reference QA}
\label{sec:supp-real-qa}

The real-video extension replaces exact simulator body state with
HMR2.0b~\citep{fourdhumans} pseudo-SMPL estimates. Before a track is used for body-control
training, its fitted body passes a separate keypoint-based quality check
with Keypoint R-CNN. Of $146$ candidate tracks, $143$ are retained.

The real-image reference used for KID$_{\mathrm{VRU}}$ is kept
disjoint from generator-training video. KITTI~\citep{kitti} training windows are
drawn from the training side of the drive-disjoint split, whereas the
KID reference contains validation-side images only. The final reference
contains $1{,}453$ pedestrian crops from $672$ images, with zero shared
drives between the training windows and the KID reference.

The detector evaluation is separated similarly. No real frame used for
generator training belongs to the detector validation partition.
Generator-training frames may belong to the detector's real training
partition, but the downstream measurements are always computed on
unseen real validation images.

The KITTI component of the real-video supervision also differs
temporally and spatially from the simulator clips. KITTI source video
is recorded at $10$\,Hz and placed in the $16$\,fps training clip
format without synthesizing intermediate frames. Interpolation is
avoided because an interpolated image would not have a corresponding
pseudo-SMPL estimate. The fixed VRU crops from this source are also
upscaled by approximately $1.9\times$ before entering the training
pipeline. These acquisition differences apply specifically to the
KITTI portion of the real-video data and should not be generalized to
the Argoverse~2~\citep{argoverse2} component.

These controls do not make pseudo-labelled real supervision equivalent
to exact simulator supervision. Rather, they ensure that the
appearance results are not explained by train--reference leakage and
that the retained real tracks pass a separate body-pose quality check.

\section{Implementation and Reproducibility}
\label{sec:supp-implementation}

The main paper specifies the Grounded Body Layer construction, control
architecture, training objective, and inference composition. Here, we
record only the additional implementation details needed to reproduce
the executed pipeline and the provenance used to maintain paired
comparisons.

\subsection{Conditioning-Layer Reproducibility Notes}
\label{sec:supp-gbl-construction}

The mathematical construction of the Grounded Body Layers is given in
Main Paper Sec.~3.1. The canonical $x$ and $z$ normalization uses
extrema from the fixed $\beta=0$ template; these coordinates are
therefore defined in one common rest-pose system rather than
renormalized per actor or frame. The third correspondence channel is
different: its extrema are computed from the visible registered body of
the corresponding actor in each frame, so it represents intra-body
depth ordering rather than absolute scene distance.

Body registration is performed independently for each actor--frame
pair. If the required body-pose representation is unavailable, only
that actor--frame pair falls back to the 2D-skeleton conditioning
representation. Other VRUs in the frame are unchanged, and fallback
does not modify the simulator-derived evaluation skeleton or GT box.
Representation coverage and registration quality are quantified in
Supplementary Sec.~S2.7.

\subsection{Training and Inference Details}
\label{sec:supp-training-details}

Main Paper Sec.~4.2 reports the control-branch architecture,
initialization, training duration, VRU weighting, composition weights,
and common inference prompt. Post-training follows the released Cosmos
rectified-flow configuration~\citep{cosmostransfer25} with logit-normal flow-time sampling
(shift $5$), while the timestep-embedder path is kept in FP32.
Training clips contain $21$ frames and map to temporal latent length
$6$. Exponential moving average is disabled, and all reported
evaluations use the raw post-training checkpoint.

The correspondence control video follows the released Cosmos control
input pathway: it is encoded into the video latent representation and
mapped to the corresponding spatiotemporal token grid. Text
conditioning is empty during control-branch post-training. At
generation time, the common descriptive driving prompt specified in the
main paper is used for every compared method. Evaluation clips contain
$93$ frames and use temporal latent length $24$.

\subsection{Compute and Artifact Provenance}
\label{sec:supp-compute}

Control-branch post-training is performed on NVIDIA A100 GPUs. Training
one control branch for the post-training schedule of Supplementary
Sec.~S5.2 requires approximately $24$ GPU-hours.

Each generated clip is linked to its conditioning specification,
generator code revision, checkpoint, and inference seed. A master
manifest indexes these fields together with the corresponding
clip$\times$configuration output. The same clip identifier links the
simulator recording, projected GT, conditioning layers, and generated
video.

Generated videos are reused for pose,
identity, temporal, appearance, and downstream analyses rather than
regenerated separately for each metric. The exported simulator source
and evaluation reference are likewise reused across generation
configurations. Consequently, paired comparisons share the source
scene, articulated motion, camera trajectory, and simulator-derived
reference; only the generated output changes with the evaluated
configuration.

\end{document}